\documentclass[journal]{IEEEtran}
\usepackage{cite}

\ifCLASSINFOpdf
  \usepackage[pdftex]{graphicx}
\else
\fi
\usepackage{amsmath}
\usepackage{amsfonts}
\usepackage{mathptmx}
\usepackage{helvet}
\usepackage{array}
\usepackage{url}

\usepackage{graphicx}

\usepackage[colorlinks,
            linkcolor=red,
            anchorcolor=green,
            citecolor=blue
            ]{hyperref}
            
\usepackage{subfig}

\DeclareMathAlphabet{\mathpzc}{OT1}{pzc}{m}{it}

\begin{document}
%
\title{Deep Learning from Noisy Image Labels with Quality Embedding}
%
%
%
%


\author{Jiangchao~Yao,~Jiajie~Wang,~
    Ivor~Tsang,~Ya~Zhang,~Jun~Sun,~
    Chengqi~Zhang,~Rui~Zhang%
    \thanks{Jiangchao Yao is with both the Cooperative Medianet Innovation
Center, Shanghai Jiao Tong University and Center for Artificial Intelligence, FEIT, University of Technology Sydney. Email:
Sunarker@sjtu.edu.cn or Jiangchao.Yao@student.uts.edu.au}%
    \thanks{Jiajie Wang and Ya Zhang are with the Cooperative Medianet Innovation
Center, Shanghai Jiao Tong University, Shanghai 200240, China. E-mail:
\{ww1024,ya\_zhang\}@sjtu.edu.cn}
    \thanks{Ivor Tsang and Chengqi Zhang are with Center for Artificial Intelligence, FEIT, University of Technology Sydney, Ultimo, NSW 2007, Australia. E-mail: \{ivor.tsang,chengqi.zhang\}@uts.edu.au.}
    \thanks{Jun Sun and Rui Zhang are with the Institute of Image Communication and
Network Engineering, Shanghai Jiao Tong University, Shanghai 200240,
China. E-mail: \{junsun,zhang\_rui\}@sjtu.edu.cn}
    }

%
%

\markboth{Journal of \LaTeX\ Class Files,~Vol.~14, No.~8, August~2015}%
{Shell \MakeLowercase{\textit{et al.}}: Bare Demo of IEEEtran.cls for Computer Society Journals}
%



\IEEEtitleabstractindextext{%
\begin{abstract}
There is an emerging trend to leverage noisy image datasets in many visual recognition tasks. However, the label noise among the datasets severely degenerates the \mbox{performance of deep} learning approaches. Recently, one mainstream is to introduce the latent label to handle label noise, which has shown promising improvement in the network designs. Nevertheless, the mismatch between latent labels and noisy labels still affects the predictions in such methods. To address this issue, we propose a quality embedding model, which explicitly introduces a quality variable to represent the trustworthiness of noisy labels. 
Our key idea is to identify the mismatch between the latent and noisy labels by embedding the quality variables into different subspaces, which effectively minimizes the noise effect. At the same time, the high-quality labels is still able to be applied for training. 
To instantiate the model, we further propose a Contrastive-Additive Noise network (CAN), which consists of two important layers: (1) the contrastive layer estimates the quality variable in the embedding space to reduce noise effect; and (2) the additive layer aggregates the prior predictions and noisy labels as the posterior to train the classifier. Moreover, to tackle the optimization difficulty, we deduce an SGD algorithm with the reparameterization tricks, which makes our method scalable to big data. 
We conduct the experimental evaluation of the proposed method over a range of noisy image datasets. Comprehensive results have demonstrated CAN outperforms the state-of-the-art deep learning approaches.
\end{abstract}

}

\maketitle

\IEEEdisplaynontitleabstractindextext

%
\IEEEpeerreviewmaketitle

\section{Introduction}

%
%
%
%
\IEEEPARstart{W}{hile} editorially labeled image data is crucial to visual classification \cite{NIPS2012_4824,simonyan2014very,Szegedy_2015_CVPR,he2016deep}, weakly supervised detection and segmentation \cite{wang2014weakly,bilen2016weakly,wang2014joint,Zhang_2015_CVPR,Khoreva_2017_CVPR,lu2017learning}, collecting such datasets in large volume can be prohibitive. Non-editorial means such as social tagging and crowdsourcing, have been explored as efficient alternatives \cite{Divvala_2014_CVPR,chen2015webly,krishna2017visual}. For example, there are a plethora of images with tags available on the Flickr website, which provides us valuable labeled resources to build image classifiers. However, the challenges lie in the fact that social tags as labels are highly noisy. As a result, deep learning from noisy image labels has attracted the increasing attention \cite{Sukhbaatar2015Training}.

Previous studies have investigated the label noise  \cite{raykar2010learning,natarajan2013learning,Liu2014Classification,zhou2012learning,Frenay2014classification} for non-deep approaches in the machine learning community. For example, Vikas \textit{et al.} \cite{raykar2010learning} introduce parameters for annotators to transit latent predictions to noisy labels. For parameter estimation, they resort to an EM optimization algorithm that is also adopted in the contemporaneous works. However, it is not straightforward to apply these studies to deep learning methods due to the computational consuming in the EM optimization.

With the success of deep learning in computer vision \cite{NIPS2012_4824,simonyan2014very,Szegedy_2015_CVPR,he2016deep}, training neural network with noisy image labels has also been explored \cite{Mnih2012Learning,Sukhbaatar2015Training,azadi2015auxiliary,misra2016seeing,Izadinia2015Deep,reed2014training,patrini2016making,xiao2015learning,jindal2016learning,li2017learning,veit17learning}. These methods can be summarized into two categories, building the robust loss function and modeling the latent label. The former paradigm is heuristic and usually depends on non-trivial hyperparameter selection. For instance, Reed \textit{et al.} \cite{reed2014training} construct a weighted combination of noisy image labels and predictions to supervise the network training. However, it is unclear that how the weight interacts with the real-world label noise for settings. 
One popular example of the latter paradigm, Sukhbaatar \textit{et al.} \cite{Sukhbaatar2015Training} model the latent label to handle the label noise. Specifically, the classifier is trained based on latent labels, and thus the label noise will not directly affect the classifier. However, they adapt latent labels to noisy labels with a linear transition layer, which cannot sufficiently model the label corruption. Label noise can still go through this layer to degenerate the performance.  
The deficiency of above deep learning methods is that they do not explicitly model the trustworthiness of noisy labels. Implicitly considering noise in the loss function or by modeling the latent label may harm the nature of noise, e.g., flip and outlier. 

\begin{figure*}[ht]
    \centering
    \includegraphics[width=160mm]{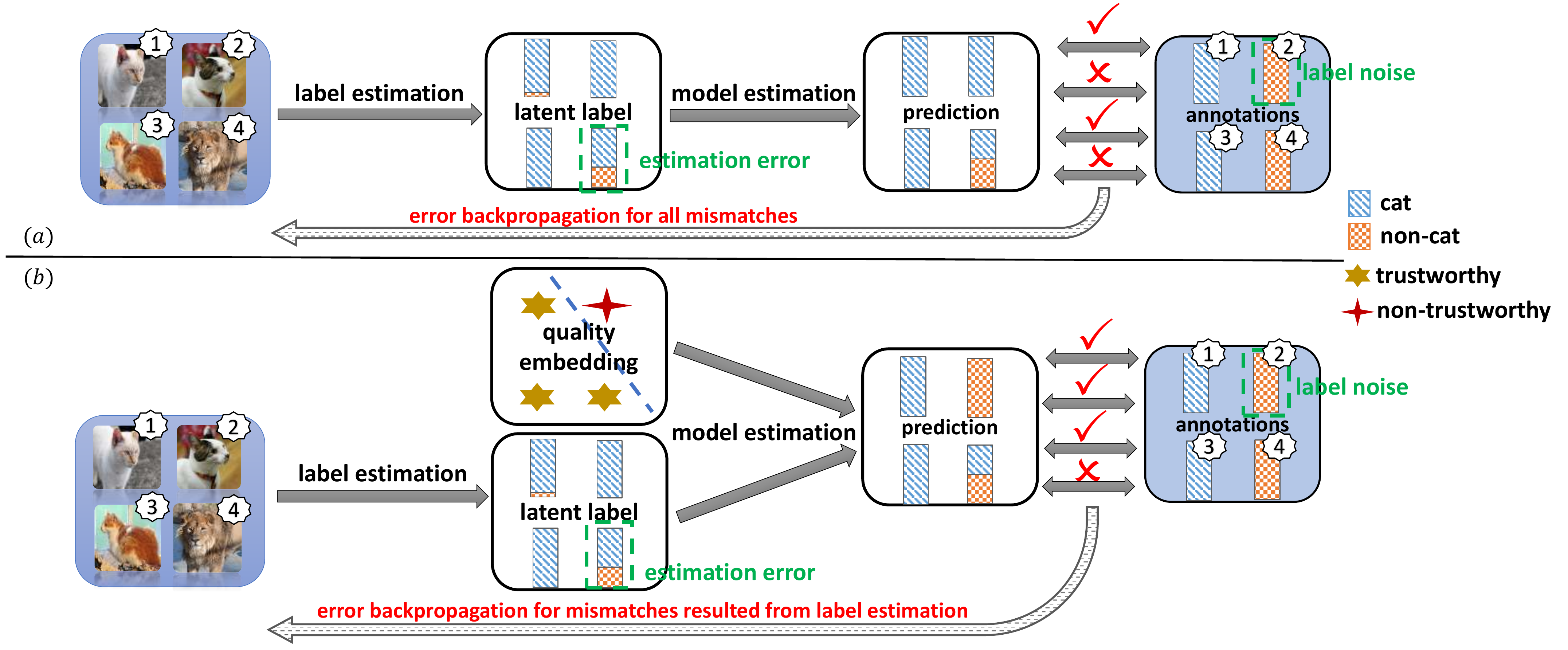}
    \caption{\small Analysis about back-propagation in previous methods that model the latent label, as well as our idea to avoid the effect of label noise. (a) All images are forward into the model and the mismatch error caused by both label estimation and label noise are back-propagated. (b) With quality embedding as a control from latent labels to predictions, the negative effect of label noise is reduced in the back-propagation. }
    \label{fig:motivation}
\end{figure*}

In this paper, we follow the latter paradigm and propose a quality embedding model. Fig. \ref{fig:motivation} illustrates our idea as well as its advantage to reduce the noise effect.
For example, in Fig. \ref{fig:motivation}(a), the latent labels and predictions of the first three cat images must approximately consistent due to their content similarity. However, mismatch will occur between the second prediction and the corresponding annotation by virtue of the label noise. For the fourth image, the prediction induced by the estimation error of the latent label, also has conflict with the fourth annotation. As a result, these two mismatches will mix together for back-propagation. On the other hand, if we explicitly introduce a quality variable to model the trustworthiness of noisy labels like Fig. \ref{fig:motivation}(b), label noise can be reduced more effectively. For example, if the quality variable of the second sample is embedded in the non-trustworthy subspace, the latent label can be disturbed accordingly to prevent mismatch error caused by the label noise from back-propagation. While for the fourth sample whose quality variable is estimated in the trustworthy subspace, the latent label still transits to the final prediction causing the mismatch. Then supervision from the correct annotations is normally fed back. 

Mathematically, we illustrate the corresponding graphical model in Fig. \ref{fig:model}. Different from previous latent-label-based deep learning approaches, a quality variable is specially introduced to model the trustworthiness of noisy labels. By embedding the quality variable into different subspaces, the shortcoming illustrated like Fig.\ref{fig:motivation}(a) can be solved as Fig.\ref{fig:motivation}(b). To instantiate our probabilistic model with deep neural network, we further design a Contrastive-Additive Noise network (CAN) shown in Fig. \ref{fig:network}. For parameter learning, we optimize an evidence lower bound \cite{wainwright2008graphical,ICML2012Paisley_687,blei2017variational} plus a variational mutual information regularizer, and deduce an SGD algorithm. The major contribution in this paper can be summarized into four parts in the following.
\begin{itemize}
    \item To address the shortcoming of existing latent-label-based deep learning approaches, we propose a quality embedding model that introduce a quality variable to represent the trustworthiness of noisy labels. By embedding the quality variable into different subspace, the negative effect of label noise can be effectively reduced. Simultaneously, the supervision from high quality labels still can be back-propagated normally for training.
    \item To instantiate the quality embedding model, we design a Contrastive-Additive Noise network. Specially, it consists of two important layers: (1) the contrastive layer estimates the quality variable in the embedding space to reduce noise effect; (2) the additive layer aggregates prior predictions and noisy labels as posterior to train the classifier.
    \item  To tackle the optimization difficulty, we apply the reparameterization tricks and deduce an efficient SGD algorithm, which makes our model scalable to big data.
    \item We conduct a range of experiments to demonstrate that CAN outperforms existing state-of-the-art deep learning methods on noisy datasets. We further present qualitative analysis about quality embedding, latent label estimation and noise pattern to give a deep insight on our model. 
\end{itemize}

The rest of this paper is organized as follows. Section 2 briefly reviews the related work of learning with noisy labels in deep learning. Then we introduce our quality embedding model, the corresponding instantiation Contrastive-Additive-Noise network as well as its optimization algorithm in Section 3. We validate the efficiency of our method over a range of experiments in Section 4. Section 5 concludes the paper.

\section{Related Work}
Social websites and crowdsourcing platforms provide us an effective way to gather a large amount of low-cost annotations for images. However, in the visual recognition tasks such as image classification, the noise among labels shall severely degenerate the performance of classification models \cite{nettleton2010}. To exploit the great value of noisy labels, several noise-aware deep learning methods have been proposed for the image classification task. Here, we briefly review these related works.

\textbf{Robust loss function} This line of research aims at designing a robust loss function to alleviate noise effect. For instance, Joulin \textit{et al.} \cite{Joulin2015Learning} weight the cross-entropy loss with the sample number to balance the emphasis of noise in positive and negative instances. Izadinia \textit{et al.} \cite{Izadinia2015Deep} estimate a global ratio of positive samples to weaken the supervision in the loss function. Reed \textit{et al.} \cite{reed2014training} consider the consistency of predictions in similar images and apply bootstrap to the loss function. They substitute the noisy label with a weight combination of the noisy label and the prediction to encourage the consistent output. \mbox{ Recently, Li \textit{et al.} \cite{li2017learning}} re-weight the noisy label with a soft label learned from side information. They train a teacher network with the clean dataset to compute the soft label by leveraging the knowledge graph. The soft label is then combined with the noisy label in the loss function to pilot student model's learning. Andreas \textit{et al.} \cite{veit17learning} rectify labels in the cross-entropy loss with a label-correction network trained on the extra clean dataset. While these methods are concerned with modifying the labels in the loss function by re-weighting or rectification, our approach also models the auxiliary trustworthiness of noisy image labels to reduce the noise effect on training.

\textbf{Modeling the latent labels} This paradigm targets at modeling the latent labels to train the classifier, and building a transition for adaption from the latent labels to the noisy labels. With the success of deep learning in image recognition, this kind of idea receives considerable attention. Mnih \textit{et al.} \cite{Mnih2012Learning} first propose a latent variable model on aerial images, which assumes that the noise is symmetric and at random. Based on it, \cite{Sukhbaatar2015Training,jindal2016learning} use an linear adaptation layer to model the asymmetric label noise, and add the layer on top of a deep neural network. This transition layer can be deemed as the confusion matrix representing label flip probability. However, the matrix only depends on the distribution of labels but ignore the information of image contents. Chen \text{et al.} \cite{chen2015webly} apply a two-stage approach to model the latent label and learn the translation to the noisy label, in which a clean dataset is  used. Different from methods that model label transition in the dataset level, Xiao \textit{et al.} \cite{xiao2015learning} propose a probabilistic graphic model that disturbs the label in the image level. However, the model also needs a small part of clean data to learn conditional probability, which may constrains the generalization of the model. To demonstrate the human-centric noisy label exhibits specific structure that can be modeled, Misra \textit{et al.} \cite{misra2016seeing} build two parallel classifiers. One classifier deals with image recognition and the other classifier model human's reporting bias. However, it still suffers from the problem mentioned in Fig. \ref{fig:motivation}(a) since similar images have similar latent variables. Although these methods take advantages of deep neural network to model the latent label, the simple transition cannot sufficiently model the label corruption. We go on by unearthing the annotation quality from training data and further utilize it to guide the learning of our model.

\section{Quality Embedding models}
\subsection{Preliminaries}
Consider that we have a noisy image dataset of $M$ items,  $$\mathpzc{D}:\left\{(x_1,y_1),(x_2,y_2),...,(x_M,y_M)\right\},$$ where each tuple in the dataset consists of one image $x_m$ and its noisy labels $y_m$. Note that $x_m$ can be the original image or the feature vector extracted from the image. $y_m\in \mathbb{R}^K$ is a $K$-dimensional binary vector indicating which labels are annotated, and $K$ is the number of categories. However, $y_m$ may be corrupted with annotation noise and thus incorrect.  We assume the underlying clean label is $z_m\in \mathbb{R}^K$. We introduce $s_m$, a quality variable embedded in $D$-dimensional Gaussian space, to represent the annotation quality of $y_m$.  For ease of reference, we list the notations of this paper in Table \ref{tab:notations}. 

Formally, it is a multi-label, multi-class classification problem with noise in labels. We target to train a deep classifier from these noisy training samples. There are many other tasks that are consistent with this setting, like weakly supervised object detection and segmentation \cite{wang2014weakly,bilen2016weakly,wang2014joint,Zhang_2015_CVPR,Khoreva_2017_CVPR,lu2017learning} with web data.

\begin{figure}[t]
    \centering
    \includegraphics[width=60mm]{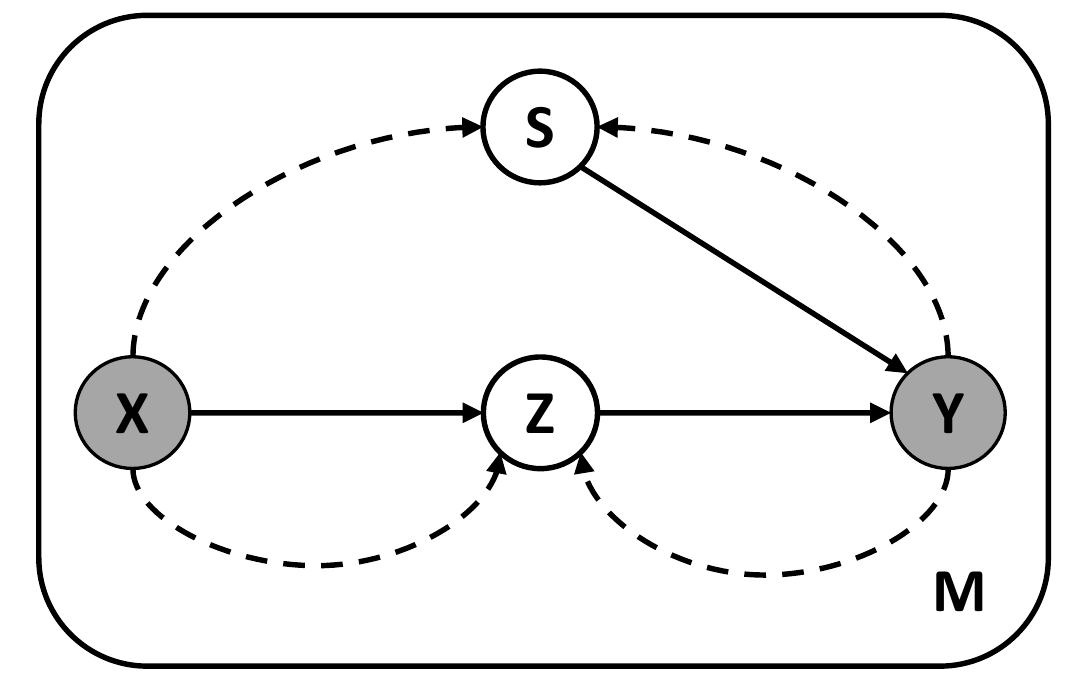}
    \caption{\small Quality embedding model for noisy image labels. The shaded nodes as the observed variables are image X and its noisy label vector Y. The latent label vector Z and the quality variable vector S are latent variables. Solid lines and dashed lines represent the generative process and the inference process respectively.}
    \label{fig:model}
\end{figure}

\subsection{Quality Embedding Model}
\subsubsection{quality embedding}
In this section, we introduce a quality variable in parallel to the latent label, which jointly transit to the noisy image label. Our probabilistic graphical model is illustrated in Fig. \ref{fig:model}. In the generative process, the latent label vector $Z$ purely depends on the instance $X$. We model this dependency with $P(Z|X)$. However, the noisy label vector $Y$ is generated based on both the annotation quality $S$ and the latent label $Z$, which we model with $P(Y|S,Z)$. In the inference process, both the distributions of $Z$ and $S$ are all modeled based on $X$ and $Y$. We respectively represent these two distributions with $q(Z|X,Y)$ and $q(S|X,Y)$, which plays roles of posterior approximation.

According to the graphical model in Fig. \ref{fig:model}, once given the training set, we have the following log-likelihood.
\begin{equation}\label{model}
\begin{split}
\ln P(Y|X) & = \sum_{m=1}^M \ln P(y_m|x_m) \\
& = \sum_{m=1}^M \ln{\int_{s_m} \sum_{z_m} P(y_m|z_m,s_m) P(z_m|x_m) P(s_m) d{s_m} }\\
& = \sum_{m=1}^{M} \ln \mathbf{E}_{P(z_m|x_m), P(s_m)}\left[P(y_m|z_m,s_m)\right] \\
\end{split}
\end{equation}
However, the log-likelihood function is difficult to explicitly compute. We instead choose to optimize an adjustable evidence lower bound (ELBO) \cite{wainwright2008graphical,ICML2012Paisley_687,blei2017variational}. The ELBO is acquired by introducing two variational distributions $q(z_m|x_m,y_m)$  and $q(s_m|x_m,y_m)$ to approximate the true distributions of $z_m$ and $s_m$. We illustrate the form of our ELBO in Eq.\eqref{ELBO}.
\begin{align} \label{ELBO}
\begin{split}
\ln{P(Y|X)}
& \geq \sum_{m=1}^{M} \mathbf{E}_{q(z_m|x_m,y_m), q(s_m|x_m,y_m)}\left[\ln P(y_m|z_m,s_m)\right] \\
&\quad\quad -\sum_{m=1}^{M} \mathbf{D_{KL}}\left[q(z_m|x_m,y_m)||P(z_m|x_m)\right] \\
&\quad\quad -\sum_{m=1}^M \mathbf{D_{KL}}\left[ q(s_m|x_m,y_m)|| P(s_m)\right]\\
\end{split}
\end{align}
Above bound is a good approximation of the marginal likelihood,
which provides a basis for selecting a model \cite{blei2017variational}. When the gap between marginal likelihood and ELBO becomes zero, the variational distributions approach the true distributions.

\begin{table}[t]
\small
\centering
\caption{Notations and their descriptions frequently used in this paper}\label{tab:notations}
\begin{tabular}{l|l}
\hline
Notation & Description \\
\hline
$M$ & number of training items \\
$K$ & number of categories \\
$D$ & dimension of the quality variable \\
$X$ & image variable \\
$Y$ & noisy label vector variable\\
$Z$ & latent label vector variable \\
$S$ & quality vector variable \\
$W_C$ & parameter of classifier network \\
$W_N$ & parameter of noise network \\
$W_Q$ & parameter of annotation quality network \\
$W_L$ & parameter of latent label network \\
$m$ & index of an item \\
$x_m$ & $m$th observed image \\
$y_m$ & $m$th observed noisy label vector \\
$z_m$ & $m$th latent label vector \\
$s_m$ & $m$th quality vector \\
$\mu$ & mean of Guassian distribution \\
$\sigma$ & covariance diagonal of Gaussian distribution \qquad\\
$\lambda$ & regularizaion cofficient\\
$\gamma_m$ & $m$th sample from Gumbel distribution \\
$\zeta_m$ & $m$th sample from Gaussian distribution \\
$\tau$ & temperature in Gumbel-SoftMax \\
$\alpha$ & time-varying coefficient \\
\hline
\end{tabular}
\end{table}

\subsubsection{variational mutual information regularizer}
Although Fig. \ref{fig:model} presents the structure prior of our probabilistic model, optimization on ELBO may not converge to the desirable optimal since modeling the distribution with neural network introduces much flexibility. It is a common problem in Bayesian models and a general solution is posterior regularizations \cite{ganchev2010posterior}. Posterior regularizations ensure the desirable expectation and simultaneously retain the computational efficiency. Such methods have been applied in clustering \cite{NIPS2010_4154}, classification \cite{zhu2014bayesian} and image generation\cite{NIPS2016_6399}. In this paper, we introduce the regularization for variational distributions of $Z$ and $S$ in the perspective of mutual information maximization. We deduce the regularizers as follows, 
\begin{align} \label{MIRZ}
\nonumber
& I((Z,S);(X,Y)) \\ \nonumber
& = H(Z,S) - H(Z,S|X,Y) \\ \nonumber
& = \mathbf{E}_{q(X,Y)}\left[\mathbf{E}_{q(Z,S|X,Y)}\left[\ln{q(Z,S|X,Y)}\right]\right] + const\\ 
&\simeq \frac{1}{M}\sum_{m=1}^M \mathbf{E}_{q(z_m|x_m,y_m)}\left[\ln{q(z_m|x_m,y_m)}\right]  \\ \nonumber
&\quad + \frac{1}{M}\sum_{m=1}^M \mathbf{E}_{q(s_m|x_m,y_m)}\left[ \ln{q(s_m|x_m,y_m)}\right] + const , \nonumber
\end{align}
where $I(;)$ means the mutual information of two distributions and $H(\cdot)$ is the entropy of the variable. As can be seen in Eq. \eqref{MIRZ}, maximizing the mutual information is equal to minimizing the entropy of $z_m$ and $s_m$. For the latent label $z_m$, such posterior regularization can force the probability $q(z_m|x_m,y_m)$ close to the extreme points. And for the quality variable $s_m$, it will encourage the distribution $q(s_m|x_m,y_m)$ to have a low variance.

\subsubsection{objective}
Combining Eq. \eqref{ELBO} with \eqref{MIRZ}, our objective then becomes the maximization of ELBO along with the mutual information regularizer. Note that, we substitute $\frac{1}{M}$ in Eq. \eqref{MIRZ} with a coefficient $\lambda\in[0,+\infty]$ to weight the regularization effect in the optimization. Instead of maximization, we re-write our goal as the following minimization problem for the simplicity sake.
\begin{align} \label{objective}
\begin{split}
\min{\text{ }\hat{L}}
& =- \sum_{m=1}^{M} \mathbf{E}_{q(z_m|x_m,y_m), q(s_m|x_m,y_m)}\left[\ln P(y_m|z_m,s_m)\right] \\
&\quad + \sum_{m=1}^{M} \mathbf{D_{KL}}\left[q(z_m|x_m,y_m)||P(z_m|x_m)\right] \\
&\quad + \sum_{m=1}^M \mathbf{D_{KL}}\left[ q(s_m|x_m,y_m)|| P(s_m)\right]\\
&\quad - \lambda \sum_{m=1}^M \mathbf{E}_{q(z_m|x_m,y_m)}\left[\ln{q(z_m|x_m,y_m)}\right] \\
&\quad - \lambda \sum_{m=1}^M \mathbf{E}_{q(s_m|x_m,y_m)}\left[\ln{q(s_m|x_m,y_m)}\right]\\
\end{split}
\end{align}
From Eq. \eqref{objective}, our model mainly differs from previous methods in three aspects. First, $P(y_m|z_m,s_m)$ indicates that the transition from the latent label to the noisy label is based on both $z_m$ and $s_m$ while previous methods \cite{Mnih2012Learning,Sukhbaatar2015Training,jindal2016learning} only depend on $z_m$. Second, previous works \cite{Mnih2012Learning,Sukhbaatar2015Training,jindal2016learning,xiao2015learning,misra2016seeing} use the linear transition $P(y_m|z_m)$ while our model applies nonlinear implementation $P(y_m|z_m,s_m)$. Third, $z_m$ and $s_m$ are approximated with $q(z_m|x_m,y_m)$ and $q(s_m|x_m,y_m)$ in the posterior perspective while previous works \cite{xiao2015learning,li2017learning,veit17learning} might have to facilitate the extra clean dataset or other label knowledge. 

\begin{figure*}
    \centering
    \includegraphics[width=110mm]{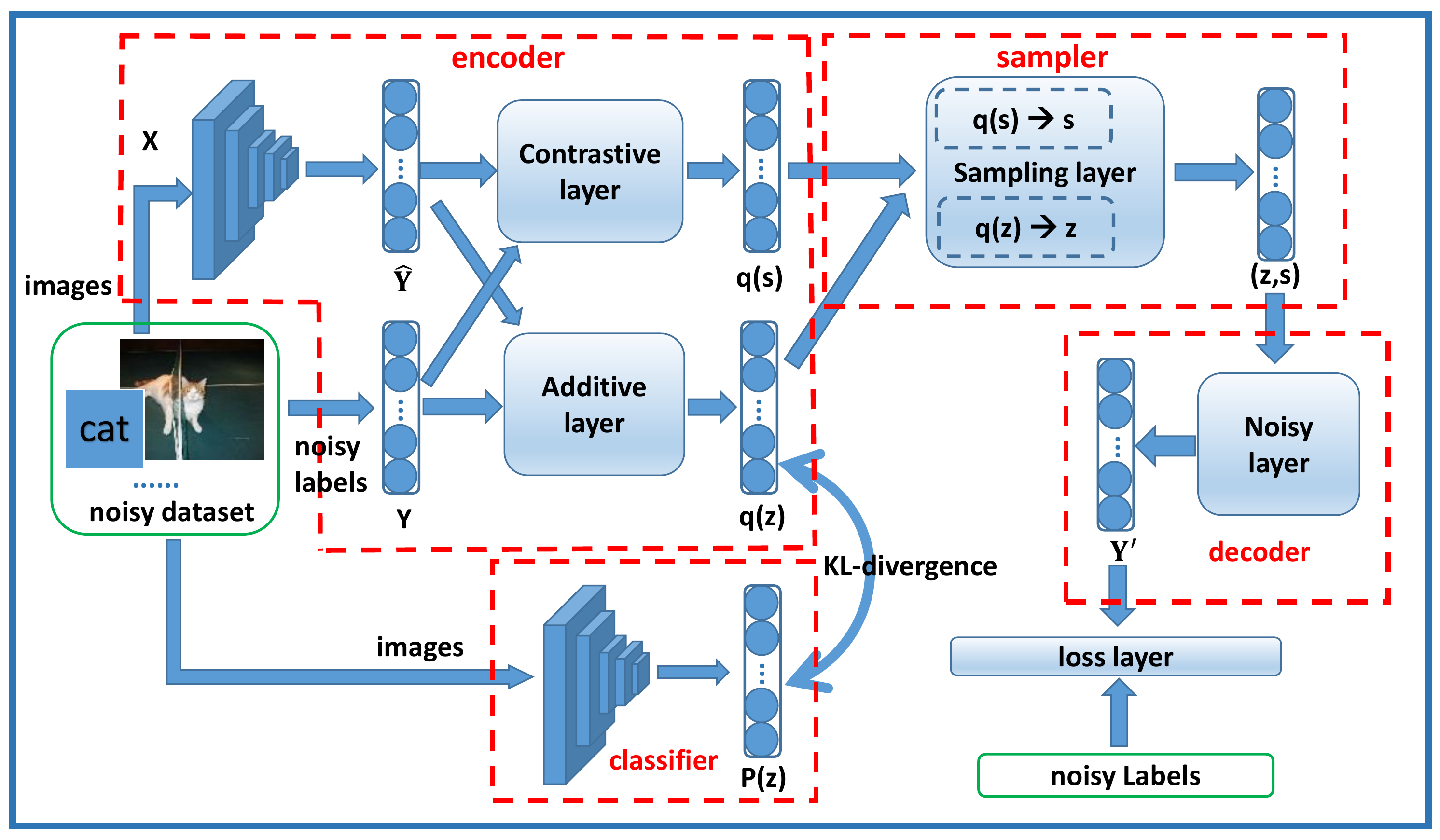}
    \caption{\small The network consists of four modules, encoder, sampler, decoder and classifier, which are trained end-to-endly. Encoder tries to learn latent labels and evaluate the quality of noisy labels; sampler is used to generate samples from  encoder outputs; decoder tries to recover noisy labels from samples. Meanwhile, our classifier is learned based on KL-divergence between $q(z)$ and $P(z)$.}
    \label{fig:network}
\end{figure*}

\subsection{Contrastive-Additive Noise Network}
In this section, we instantiate our model with a Contrastive-Additive Noise network (CAN) in Fig. \ref{fig:network}. Simply, CAN consists of four modules, encoder, sampler, decoder and classifier, which are corresponding to the different parts of our model respectively. In the following, we decribe the design in detail.

\subsubsection{architectures}For {\bf encoder module}, it is used to model the variational distributions, $q(s_m|x_m,y_m)$ and $q(z_m|x_m,y_m)$. Concretely, we first forward $x_m$ to a neural network to generate a prior label judgement $\hat{y}_m$. Then, according to $\hat{y}_m$ and $y_m$, we model the distribution parameters with two elaborately-designed layers. The neural network for $\hat{y}_m$ can be decided by the type of $x_m$. If $x_m$ is the original image, then a convolutional neural network can be applied. While if $x_m$ is a feature vector, a fully-connected network can be chosen. In Fig. \ref{fig:network}, we take the convolutional neural network as an example. The {\bf sampler module} is the implementation of Monte Carlo sampling for $q(s_m|x_m,y_m)$ and $q(z_m|x_m,y_m)$. It receives the output of the encoder module and samples from the Gumbel and Gaussian distributions to generate a sample set of $z_m$ and $s_m$. In the next section, we will talk out this part in detail with reparameterization tricks. For the {\bf decoder module}, it is a neural network for $P(y_m|z_m,s_m)$, which consists of two group of (linear, ReLU) layers, following with a Sigmoid layer. It takes the sampler output to recover noisy labels.
Previous works \cite{Mnih2012Learning,Sukhbaatar2015Training,jindal2016learning,xiao2015learning,misra2016seeing} usually use a linear transition from $z_m$ to $y_m$. We consider the nonlinear transition since we have the heterogeneous quality variable $s_m$. The {\bf classifier module} as our most important target $P(z_m|x_m)$, employs a same network for $\hat{y}_m$ in the encoder module. It is trained based on KL-divergence between $q(z_m|x_m,y_m)$ and $P(z_m|x_m,y_m)$.

\subsubsection{contrastive layer and additive layer} We specially describe these two important layers in the encoder module. Regarding the distribution $q(s_m|x_m,y_m)$, it is a $D$-dimensional Gaussian distribution and both mean and variance need to be modeled. We exploits the {\bf contrastive layer} to implement the estimation. It internally forwards $y_m$ and $\hat{y}_m$ into a shared fully-connected layer with ReLU ($\mathbf{f_s(\cdot)}$) and transforms their difference to $\mu$ and $\log{\sigma^2}$ with another fully-connected layer (function $\mathbf{f_t(\cdot)}$). It is simply represented as follows,
\begin{align}
\small
\begin{split}
(\mu(x_m,y_m),\log{\sigma^2(x_m,y_m)}) = \mathbf{f_t}(\mathbf{f_s}(y_m)-\mathbf{f_s}(\hat{y}_m)). \nonumber
\end{split}
\end{align}
This contrastive layer is built up based on the assumption that the quality variable $s_m$ is related to the difference between $y_m$ and $\hat{y}_m$. We evaluate their difference in a latent space with $\mathbf{f_s(\cdot)}$ and decide which subspace it is embedded with $\mathbf{f_t(\cdot)}$. This embedding mechanism makes us identify the label quality explicitly and subsequently helps to reduce the noise effect in $P(y_m|z_m,s_m)$. This idea has never been proposed in previous noise-aware deep learning approaches \cite{Mnih2012Learning,Sukhbaatar2015Training,azadi2015auxiliary,misra2016seeing,Izadinia2015Deep,reed2014training,patrini2016making,xiao2015learning,jindal2016learning,li2017learning,veit17learning}. 

Regarding the distribution $q(z_m|x_m,y_m)$, it consists of $K$ Bernoulli distributions and thus $K$ probabilities need to be modeled. We design an {\bf additive layer} to learn these parameters. It internally uses two non-shared fully-connected layers ($\mathbf{f_{ns1}}$ and $\mathbf{f_{ns2}}$) to transform $y_m$ and $\hat{y}_m$ into a latent space, and then feeds their addition into another fully-connected layer plus a sigmoid function (function $\mathbf{f^{'}_{t}}$), illustrated as follows,
\begin{align}
\small
\begin{split}
q(z_{m}|x_m,y_m) = \mathbf{f^{'}_{t}}(\mathbf{f_{ns1}}(y_m)+\mathbf{f_{ns2}}(\hat{y}_m)). \nonumber
\end{split}
\end{align}
This design learns a posterior label $z_m$ from $y_m$ and $\hat{y}_m$ by a nonlinear combination with neural network. Previous methods in \cite{Joulin2015Learning,Izadinia2015Deep,reed2014training, li2017learning,veit17learning} use a weight in their lost function to linearly combine the noisy label $y_m$ with the ``soft" label from the prediction, the clean dataset or other side information. They usually need non-trivial tuning manually, while we resort to a learning procedure by neural network automatically.

The whole network can be trained end-to-endly, which will be explained in the next section. In the training, the noise effect is reduced by the branch of the quality variable, and simultaneously the posterior label is estimated by the additive layer to guarantee a more reliable training. We will demonstrate the effectiveness of our network in the experiments.

\subsection{Optimization}
In this section, we will analyze the difficulty in optimization and deduce an SGD algorithm with reparameterization tricks. 
\subsubsection{The reparameterization tricks} The first term in the RHS of Eq. \eqref{objective} has no closed form when either $q(z_m|x_m,y_m)$ or $q(s_m|x_m,y_m)$ is not conjugated with $P(y_m|z_m,s_m)$. Let alone we model these distributions with deep neural network in the paper. The general way is by the Monte Carlo sampling. However, Paisley \textit{et al.} \cite{ICML2012Paisley_687} have shown when the derivative is about $q(z_m|x_m,y_m)$ or $q(s_m|x_m,y_m)$, the sampling estimation will present high variance. In this case, a large number of samples will be required to have an accurate estimation, which may lead to the high GPU load and the computational burden. Fortunately, reparameterization tricks \cite{kingma2014auto,jang2016categorical} are explored to overcome this difficulty in the recent years. They have shown promising efficiency in discrete and continuous representation learning. Simply, the idea behind reparameterization tricks is to decouple the integral variate as one parameter-related part and another parameter-free variate. After integral by substitution, the Monte Carlo sampling on this parameter-free variate will have a small variance. According to this, we apply the reparameterization trick \cite{jang2016categorical} for discrete $z_m$ and the reparameterization trick \cite{kingma2014auto} for continuous $s_m$ as follows,
\[
	\left\{ \begin{array}{l}
			 z_{mk} = g(\gamma_{mk}) = \frac{\exp\left(\left(\ln q(z_{mk}=1|x_m,y_m) + \gamma_{mk1} \right)/\tau\right)}{\sum_{v=0}^1\exp\left(\left(\ln q(z_{mk}=v|x_m,y_m) + \gamma_{mkv} \right)/\tau\right)} \\
			 s_m = f(\zeta_m) = \mu(x_m,y_m) + \sigma^2(x_m,y_m)\odot\zeta_m
		\end{array}\right\},
\]
where $\tau$ is a temperature to control the discreteness of samples, $\gamma_{mk}\sim \mathbf{Gumbel(0,1)}$\footnote{$\gamma_{mk1}$, $\gamma_{mk2}$ are both sampled by $-\log(-\log U)$, where $U\sim$ Uniform(0,1)} and $\zeta_m \sim \mathbf{N(0,1)}$ are the parameter-free variates, $q(z_m|x_m,y_m)$, $\mu(x_m,y_m)$ and $\sigma(x_m,y_m)$ are parameter-related parts. With above reparameterization tricks, we have the following low-variance sampling estimation,
\begin{align}\label{decoder}
\begin{split}
& \mathbf{E}_{q(z_m|x_m,y_m), q(s_m|x_m,y_m)}\left[\ln P(y_m|z_m,s_m)\right] \\
&\quad \simeq \frac{1}{N} \sum_{n=1}^{N} \ln P(y_m|g(\gamma_m^{(n)}),f(\zeta_m^{(n)})) ,
\end{split}
\end{align}
where $N$ is the sample number of $\gamma_m$ and $\zeta_m$ for the $m$th image. Based on Eq. \eqref{decoder}, the first term in the RHS of Eq. \eqref{objective} can be efficiently estimated, even though we set the sample number $N$ equal to 1 in the training. 

\subsubsection{Stochastic variational gradient}
The remaining terms in the RHS of Eq. \eqref{objective} can be explicity computed. We just present their deduction in the appendix. Putting Eq. \eqref{decoder} and \eqref{rkl} (in the appendix) back to Eq. \eqref{objective}, the objective is derivable regarding parameters of all distributions. We can learn the parameter of each distribution with a SGD algorithm, even if they are all modeled with deep neural network. It is important for deep learning especially on the large datasets. Assuming $W_N$, $W_C$, $W_L$ and $W_Q$ respectively represent the parameters of $P(Y|Z,S)$, $P(Z|X)$, $q(Z|X,Y)$ and $q(S|X,Y)$, their gradients can be computed with the following equations with chain rules.
\begin{align}\label{derivatives}
\small
\begin{split}
\nabla_{W_N}\hat{L} 
& = -\sum_{m=1}^{M} \frac{1}{N}\sum_{n=1}^{N} \nabla_{W_N}\ln P(y_m|g(\gamma_m^{(n)}),f(\zeta_m^{(n)})) \\
\nabla_{W_L}\hat{L} & = -\sum_{m=1}^{M}\frac{1}{N}\sum_{n=1}^{N} \nabla_{g}P(y_m|g(\gamma_m^{(n)}),f(\zeta_m^{(n)}))\nabla_{W_L}g(\gamma_m^{(n)}) \\
&\quad + \sum_{m=1}^{M} \sum_{k=1}^K \left( \ln  \frac{q(z_{mk1}|x_m,y_m)^{1-\lambda}\left(1-P(z_{mk1}|x_m)\right)}{\left(1-q(z_{mk1}|x_m,y_m)\right)^{1-\lambda}P(z_{mk1}|x_m)} \right) \\
& \qquad \nabla_{W_L} q(z_{mk1}|x_m,y_m)\\
\nabla_{W_Q}\hat{L}
& = -\sum_{m=1}^{M}\frac{1}{N}\sum_{n=1}^{N} \nabla_{f}P(y_m|g(\gamma_m^{(n)}),f(\zeta_m^{(n)}))\nabla_{W_Q}f(\zeta_m^{(n)}) \\
&\quad + \sum_{m=1}^{M}\frac{1}{2}\nabla_{W_Q} \sum_{d=1}^{D} \left( \sigma_d^2(x_m,y_m)  - (1-\lambda)\ln \sigma_d^2(x_m,y_m)\right)\\
&\quad + \sum_{m=1}^{M}\frac{1}{2}\nabla_{W_Q}\left( \mu(x_m,y_m)^T\mu(x_m,y_m) \right ) \\
\nabla_{W_C}\hat{L} & = \sum_{m=1}^{M} \sum_{k=1}^K \frac{P(z_{mk1}|x_m) - q(z_{mk1}|x_m,y_m)}{P(z_{mk1}|x_m)(1-P(z_{mk1}|x_m))} \nabla_{W_C}P(z_{mk1}|x_m) \\
\end{split}
\end{align}
where $z_{mk1}$ is the abbreviation of $z_{mk}=1$ for the space sake. Note that, although we have above gradients for CAN, there are two undesirable problems existing in the optimization: (1) It is not easy to precisely decouple the information from back-propagation respectively for $z_m$ and $s_m$, i.e., squeeze out the clean label information for $z_m$ and leave the quality-related information to $s_m$; (2) The corresponding \mbox{label order between} $z_{m}$ and $y_{m}$ may be inconsistent in the optimization. For example, the category in first dimension of $z_m$ can be corresponding to the category in the second dimension of $y_m$. To avoid these two problems, we can asymmetrically inject auxiliary information to the optimization procedure in an annealing way, that is, substitute $\nabla_{W_C}\hat{L}$ with the following Eq.\eqref{dlabel_pseudo}.
\begin{align}\label{dlabel_pseudo}
\begin{split}
\nabla_{W_C}\hat{L}_{mod} & = \left( 1-\rho(t)\right) \nabla_{W_C}\hat{L} + \rho(t)\nabla_{W_C}\hat{L}_{temp},
\end{split}
\end{align}
where $\nabla_{W_C}\hat{L}_{temp}$ is gradient regarding the cross-entropy loss between $z_m$ and $y_m$, and $\rho(t) = \exp(-\alpha*t)\text{, } \alpha>0$ is a time-varying term. In this equation, $\nabla_{W_C}\hat{L}_{mod}$ is initially decided by $\nabla_{W_C}\hat{L}_{temp}$ and then progressively anneals to $\nabla_{W_C}\hat{L}$ with $t$ increasing. It guarantees the decoupling procedure from the back-propagation with asymmetrical constraint to $z_m$ and make the label order of $z_m$ and $y_m$ consistent in the optimization.

\begin{figure}
  \centering
  \label{vae}
  \includegraphics[width=85mm]{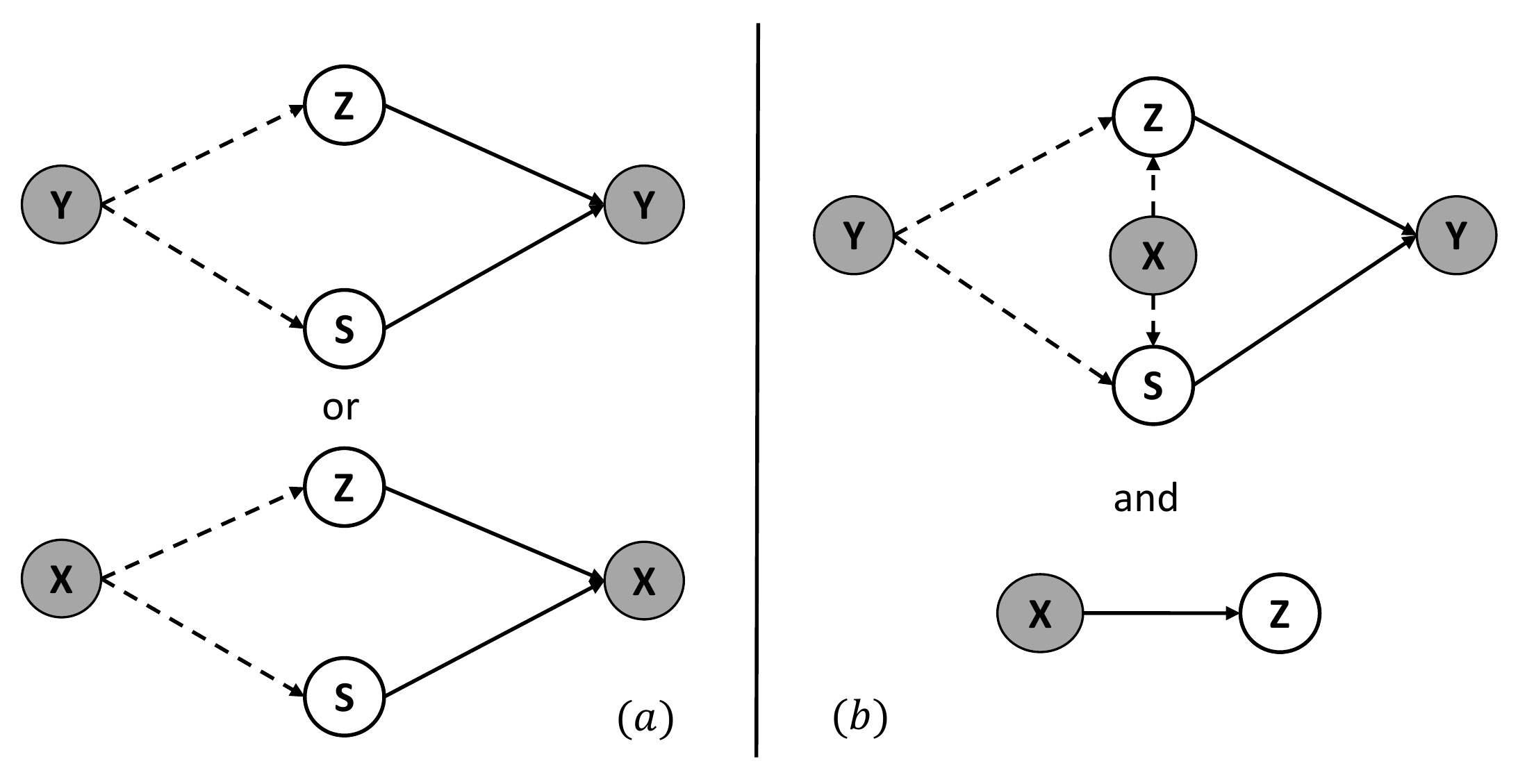}
  \caption{\small Difference between the conventional auto-encoder and our model. Solid lines and dashed lines respectively represent generative (decoding) and inference (encoding) procedures. (a) a conventional auto-encoder is symmetric that observed knowledge is used to encode to latent variables and decoded symmetrically. (b) Our model uses an auxiliary variables $X$ in this encoding-decoding procedure and meanwhile learns a discriminative part ($X$ to $Z$).}
  \label{vae}
\end{figure}

\begin{figure*}
    \centering
    \subfloat{
    \label{count}
    \includegraphics[width=97mm,height=20mm]{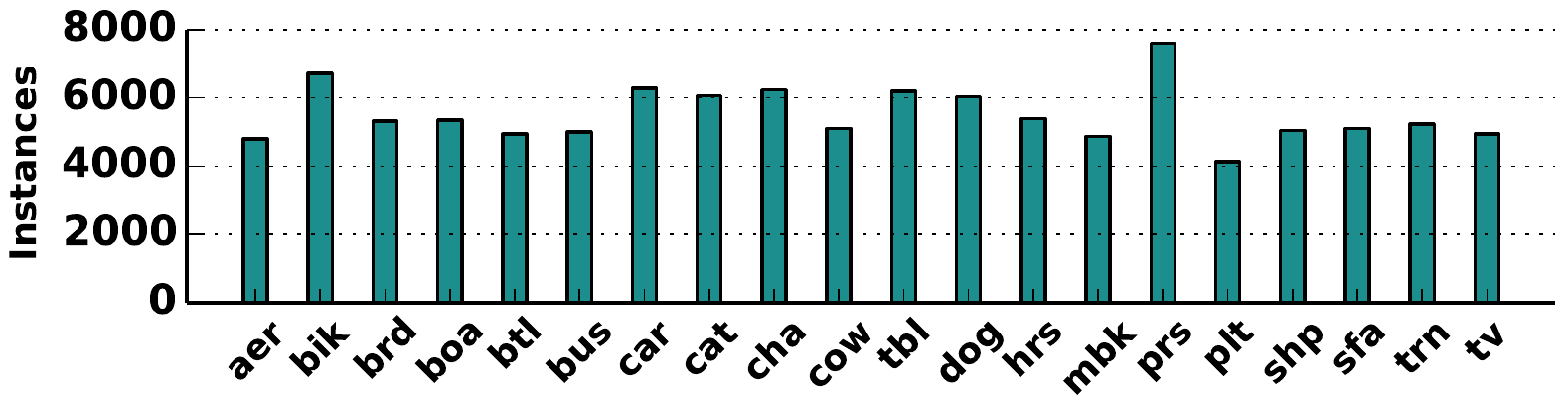}}
    \subfloat{
    \label{accuracy_fscore}
    \includegraphics[width=60mm,height=20mm]{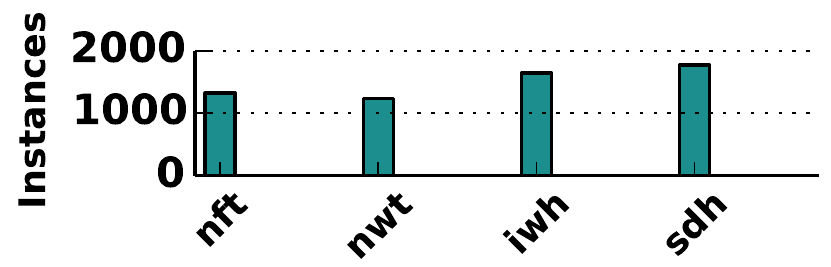}}
    \caption{\small \textbf{Left:} the instance number of each category in \emph{WEB} dataset. \textbf{Right:} the instance number of each category in \emph{AMT} dataset.}
    \label{fig:noisySet}
\end{figure*}

\begin{table*}[h]
\centering 
\caption{Classification Results on \emph{V07TE}}
\renewcommand{\arraystretch}{1.0}
\normalsize
\scalebox{0.75}{
\begin{tabular}{p{14mm} cccccccccccccccccccc c} 
\hline\hline 
Model & aer & bik & brd & boa & btl & bus & car & cat & cha & cow & tbl & dog & hrs & mbk & prs & plt & shp & sfa & trn & tv & mAP\\ [0.5ex] 
\hline
Resnet-N & 93.5 & 85.3 & 90.1 & 85.1 & 51.2 & 82.3 & 84.8 & 91.2 & 59.3 & 87.1 & 72.1 & 88.7 & 91.3 & 88.9 & 76.1 & 54.4 & 87.6 & 70.0 & 90.4 & 61.4 & 79.5\\ 
\hline
LearnQ & 92.8 & 86.1 & 91.0	& 87.8 & 50.2 & 84.9 & 85.1	& 90.9 & 59.2 & 88.3 & 71.1	& 90.1 & 91.2 & 88.1	& 78.3 & 56.6 &	89.1 & 73.1 & 90.7 & 64.3 & 80.4\\
\hline
ICNM & 92.5 & 86.2 & 90.5 &	87.9 & 47.7 & 84.0 & 84.8 &	90.6 & 59.8 & 88.3 & 72.7 &	89.8 & 91.5 &	87.2 & 77.0	& \textbf{57.0} & 88.9 &	71.5 & 91.2 & 65.7 & 80.3\\
\hline
Bootstrap & 94.0 & \textbf{88.4}	& 90.3 & 88.2 &	51.7 & 83.8	& 86.5 & 91.0 & 65.4 & 88.0	& 77.4 & 90.4 &	91.8 & \textbf{90.8} & 79.8 & 55.2 & 92.8 & 75.2 & 90.8 & 66.4 & 81.9\\
\hline
CAN & \textbf{95.5} & 87.0 & \textbf{91.4} & \textbf{89.9} & \textbf{60.1} & \textbf{85.5} & \textbf{87.6} & \textbf{92.0} & \textbf{67.2} & \textbf{90.1} & \textbf{77.7} & \textbf{91.8} & \textbf{93.3} & 90.6 & \textbf{82.1} & 56.0 & \textbf{93.6} & \textbf{80.7} & \textbf{94.5} & \textbf{70.6} & \textbf{83.8} \\
\hline
\end{tabular}}
\label{voc2007test} 
\end{table*}

\begin{table*}
\caption{Classification Results on \emph{V12TE}}
\centering 
\renewcommand{\arraystretch}{1.0}
\normalsize
\scalebox{0.75}{
\begin{tabular}{p{14mm} cccccccccccccccccccc c} 
\hline\hline 
Model & aer & bik & brd & boa & btl & bus & car & cat & cha & cow & tbl & dog & hrs & mbk & prs & plt & shp & sfa & trn & tv & mAP\\ [0.5ex] 
\hline 
Resnet-N & 98.4 & 81.1 & 92.9 & 88.7 & 57.0 & 87.4 & 73.2 & 96.6 & 63.3 & 90.0 & 63.9 & 94.3 & 95.0 & 92.9 & 76.8 & 43.8 & 92.9 & 67.2 & 93.1 & 65.1 & 80.7\\ 
\hline 
LearnQ & 98.4 & 83.8 & 93.8 & 88.5 & 53.5 & 87.8 & 73.7 & 96.5 & 64.3 & 90.6 & 62.6 & 94.6 & 96.1 & 91.6 & 78.4 & \textbf{46.8} & 92.8 & 69.0 & 94.0 & 65.4 & 81.1\\
\hline
ICNM & 98.1 & 82.9 & 93.6 & 88.9 & 53.4 & 87.7 & 72.3 & 96.2 & 64.7 & 91.2 & 66.3 & 94.2 & 96.2 & 91.4 & 78.0 & 44.0 & 93.5 & 69.3 & 94.4 & 66.9 & 81.2\\
\hline
Bootstrap & 98.6 & \textbf{84.1} & 93.6 & 90.9 & 56.3 & 89.8 & 75.5 & 96.3 & 69.8 & 91.6 & 69.9 & 94.4 & 95.8 & 93.2 & 82.2 & 43.2 & 92.8 & 70.9 & 95.4 & 67.4 & 82.6\\
\hline
CAN & \textbf{98.8} & \textbf{84.1} & \textbf{95.3} & \textbf{93.2} & \textbf{62.1} & \textbf{90.8} & \textbf{77.0} & \textbf{97.9} & \textbf{72.6} & \textbf{94.4} & \textbf{73.5} & \textbf{96.1} & \textbf{97.7} & \textbf{94.3} & \textbf{82.4} & 45.5 & \textbf{95.8} & \textbf{71.4} & \textbf{95.8} & \textbf{68.6} & \textbf{84.4}\\ 
\hline
\end{tabular}}
\label{voc2012test} 
\end{table*}

The optimization procedure can be interpreted as a probabilistic auto-encoder \cite{kingma2014auto}. However, our model is different from the traditional auto-encoder, which is illustrated in Fig.\ref{vae}. A conventional auto-encoder is symmetric, that is, observed knowledge is encoded into latent variables and decoded to itself, for instance in Fig.\ref{vae} (a), $Y$ is encoded to $Z$ and $S$, and then $Z$ and $S$ are used to decode to $Y$. It is usually used in generative models and their corresponding applications like image generation \cite{higgins2016beta,van2016conditional}. In Fig. \ref{vae} (b), our model uses an auxiliary variables $X$ in the encoding-decoding procedure, that is, $X$ and $Y$ are used to encode $Z$ and $S$, and then $Z$ and $S$ are only used to decode $Y$. Simultaneously, a discriminative model will be involved and jointly optimize with our auto-encoder.

\section{Experiments}
In this section, we conduct the quantitative and qualitative experiments to show the superiority of CAN in classification. Specifically, we compare CAN with state-of-the-art methods, investigate its performance with varying training sizes, hyperparameter sensitivity and artificial noise. To present a deep insight on how CAN works, we analyze the quality embedding, latent label estimation and noise transition in the network.

\subsection{Datasets}
We totally have five image datasets used in the experiments.

{\bf WEB\footnote{\url{https://webscope.sandbox.yahoo.com/catalog.php?datatype=i&did=67}}} This dataset is a subset of YFCC100M \cite{Thomee2015The} collected from the social image-sharing website. It is formed by randomly selecting images from YFCC100M, which belong to the 20 categories of the PASCAL VOC \cite{pascal-voc-2007}. The statistics of this dataset are shown in the left panel of Fig. \ref{fig:noisySet}. There are 97,836 samples in total and the sample number in each category ranges from 4k to 8k. Most of images in this dataset belong to one class and about 10k images have two or more. Labels in this dataset may contain annotation error.

{\bf AMT\footnote{\url{https://www.microsoft.com/en-us/research/publication/learning-from-the-wisdom-of-crowds-by-minimax-entropy/}}} This dataset is collected by Zhou \textit{et al.} \cite{zhou2012learning} from the Amazon Mechanical Turk platform. They submit 4 breeds of dog images from the Stanford Dog dataset \cite{KhoslaYaoJayadevaprakashFeiFei_FGVC2011} to Turkers and acquire their annotations. To ease the classification, Zhou \textit{et al.} also provide a 5376-dimensional feature for each image. The statistics of this dataset is illustrated in the right panel of Fig. \ref{fig:noisySet}. There are 7,354 samples in total and the sample number in each category is between 1k and 2k. All images in this dataset belong to one class. Labels in this dataset may contain annotation error.

{\bf V07\footnote{\url{http://host.robots.ox.ac.uk/pascal/VOC/voc2007/}}} This dataset is provided for the 20-cateogry classification task in PASCAL VOC Chanllenge 2007 \cite{pascal-voc-2007}. It consists of two subsets: trainging (\emph{V07TR}) and test (\emph{V07TE}). There are 5,011 samples in \emph{V07TR} and 4,592 samples in \emph{V07TE}. All labels in this dataset are clean.

{\bf V12\footnote{\url{http://host.robots.ox.ac.uk/pascal/VOC/voc2012/}}} This dataset is provided for the 20-cateogry classification task in PASCAL VOC Chanllenge 2012 \cite{pascal-voc-2012}. It consists of two subsets: trainging (\emph{V12TR}) and test (\emph{V12TE}). There are 11,540 samples in \emph{V12TR} and 10,991 samples in \emph{V12TE}. All labels in this dataset are clean.

{\bf SD4\footnote{\url{http://vision.stanford.edu/aditya86/ImageNetDogs/}}} This last dataset consists of 4 categories of dogs (same to \cite{zhou2012learning}) in the Stanford Dog dataset \cite{KhoslaYaoJayadevaprakashFeiFei_FGVC2011}. It is a fine-grained categorization dataset and there are 837 samples in total. We randomly partition samples into training (\emph{SD4TR}) and test (\emph{SD4TE}) by $3:1$ to use. All labels in this dataset are clean.

\subsection{Experimental Setup}
For \emph{WEB}, \emph{V07} and \emph{V12} datasets, a 34-layer residual network \cite{he2016deep} is adopted as the convolutional networks in CAN, and this configuration is also applied to all baselines to be fair. In the training phase, we first resize the short side of each image to 224 and then follow the transformations in the residual network\footnote{\url{https://github.com/facebook/fb.resnet.torch}} to preprocess images. In the test phase, we average the results of six-crop images as the final prediction. For \emph{AMT} and \emph{SD4} datasets, we directly use the features provided by  \cite{zhou2012learning}. Hence, one 3-layer perception network (5376$\rightarrow$1024, ReLU, 1024$\rightarrow$30, ReLU, 30$\rightarrow$4) is adopted as the substitution of the convolutional networks in CAN. Both the temperature $\tau$ in the Gumbel-softmax function and the annealing coefficient $\rho$ in Eq. \eqref{dlabel_pseudo} vary with the formula $\max\left(0.5,\exp(-3\text{x}10^{-5}\text{xStep})\right)$. $N$ in the sampler is set to 1 following \cite{kingma2014auto}. The regularizer coefficient $\lambda$ is empirically set to 0.3. The batch size is set to 50 and the learning rate starting from 0.01 is divided by 10 every 30 epochs. All experiments run 90 epochs. For the evaluation metric, we adopt Average Precision (AP) and mean Average Precision (mAP) like \cite{pascal-voc-2007,pascal-voc-2012}.

\begin{figure*}
    \centering
    \includegraphics[width=182mm]{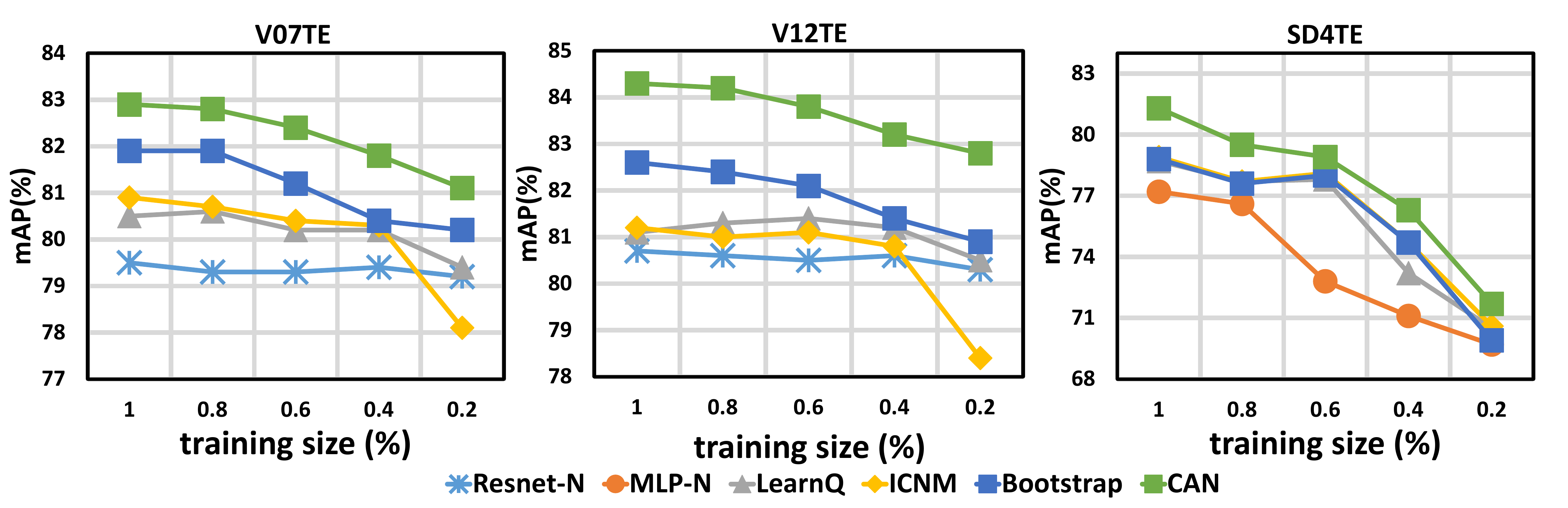}
    \caption{\small Classification results with different training sizes. We sample the subsets of \emph{WEB} and \emph{AMT} with five different ratios for training, and evaluate all models on \emph{V07TE}, \emph{V12TE} and \emph{SD4TE} datasets.}\label{fig:trainSize}
\end{figure*}

In the following sections of ``model comparision", ``impact of training size" and ``hyperparameter sensitivity", we train all models on \emph{WEB} and \emph{AMT} datasets and test them on \emph{V07TE}, \emph{V12TE} and \emph{SD4TE} datasets. Note that, models trained on \emph{WEB} dataset are evaluated on both \emph{V07TE} and \emph{V12TE} datasets since they have same categories. And models trained on \emph{AMT} dataset are ony evaluated on \emph{SD4TE} dataset. For the ``artificial noise" section, we first quantitatively add noise to \emph{V07TR}, \emph{V12TR} and \emph{SD4TR} datasets, and then train all models. Finally, we test them on \emph{V07TE}, \emph{V12TE} and \emph{SD4TE} datasets.

\subsection{Classification Results}
\subsubsection{Training with real-world noisy datasets}
To demonstrate the effectiveness of the proposed method in classification, we compare CAN with three state-of-the-art approaches, LearnQ \cite{Sukhbaatar2015Training}, ICNM \cite{misra2016seeing} and Bootstrap \cite{reed2014training}. Besides, two baselines Resnet-N and MLP-N are added, which directly train the 34-layer residual network and the 3-layer perception network on \emph{WEB} dataset and \emph{AMT} dataset. The classification performance for each category on the \emph{V07TE}, \emph{V12TE} and \emph{SD4TE} datasets is reported in Table. \ref{voc2007test}, 
\ref{voc2012test},
and \ref{dog}.

From the results in Table \ref{voc2007test} and \ref{voc2012test}, we find CAN outperforms all baselines in terms of mAP and show improvement almost in all categories. For example, on \emph{V07TE} dataset, CAN achieves 83.8$\%$ mAP, which outperforms Resnet-N by 4.3$\%$ mAP and the best baseline Bootstrap by 1.9$\%$ mAP. In the challenging categories such as ``bottle", ``chair" and ``sofa", it also achieves significant improvement. However, although the results of LearnQ, ICNM and Bootstrap are better than those of Resnet-N, the improvement is still limited. Similarly in Table. \ref{dog}, CAN outperforms the baselines by at least 2.8$\%$ mAP while LearnQ, ICNM and Bootstrap only improve about 1.6$\%$ mAP compared with MLP-N. 

\begin{table}
\caption{Classification Results on SD4TE}
\centering 
\renewcommand{\arraystretch}{1.0}
\normalsize
\scalebox{0.75}{
\begin{tabular}{c p{10mm} p{10mm} p{10mm} p{10mm} c} 
\hline\hline 
Model & nft & nwt & iwh & swh & mAP\\ [0.5ex] 
\hline
MLP-N & 78.1 & 73.2 & 80.9 & 76.5 & 77.2\\ 
\hline
LearnQ & 80.5 & 73.7 & 83.0 & 77.7 & 78.7\\
\hline
ICNM & 80.5 & 72.8 & \textbf{83.9} & 78.3 & 78.9 \\
\hline
Bootstrap & 80.7 & 72.5 & 83.7 & 78.1 & 78.8 \\
\hline 
CAN & \textbf{82.0} & \textbf{79.0} & 81.8 & \textbf{83.8} & \textbf{81.7}\\ 
\hline
\end{tabular}}
\label{dog} 
\end{table}

Based on above experiments, we have the following interpretations. (1) LearnQ and ICNM, which only introduce the latent label to handle the label noise, cannot prevent noise from degenerating the classifier sufficiently. (2) Bootstrap shares the similar idea with CAN in the aspect of estimating the posterior label for training. But its loss function uses the linear combination of predictions and noisy labels, which still cannot prevent the error back-propagation from label noise. (3) Our approach, which one one hand models the trustworthiness of noisy labels to reduce the noise effect, and on the other hand estimates the latent label in the posterior perspective to train the classifier, shows better classification performance.

\begin{table}[t]
\caption{Classification results with different $\lambda$ in CAN.}
\centering 
\renewcommand{\arraystretch}{1.0}
\normalsize
\scalebox{0.75}{
\begin{tabular}{c p{8mm} p{8mm} p{8mm} p{8mm} p{8mm} p{8mm} c} 
\hline\hline 
$\lambda$ & 0 & 0.2 & 0.5 & 1 & 2 & 5 & 10 \\ 
\hline
\emph{V07TE} & 82.9 & 83.5 & \textbf{84.8} & 83.6 & 80.7 & 78.8 & 77.0 \\ 
\hline
\emph{V12TE} & 84.3 & \textbf{85.2} & 84.1 & 83.0 & 80.8 & 78.3 & 76.6\\
\hline
\emph{SD4TE} & 78.6 & \textbf{80.7} & 80.4 & 79.9 & 76.4 & 73.9 & 71.3 \\
\hline
\end{tabular}}
\label{hyperparameter} 
\end{table}

\begin{figure*}[!ht]
    \centering
    \subfloat{
    \label{ssen1}
    \includegraphics[width=43mm,height=35mm]{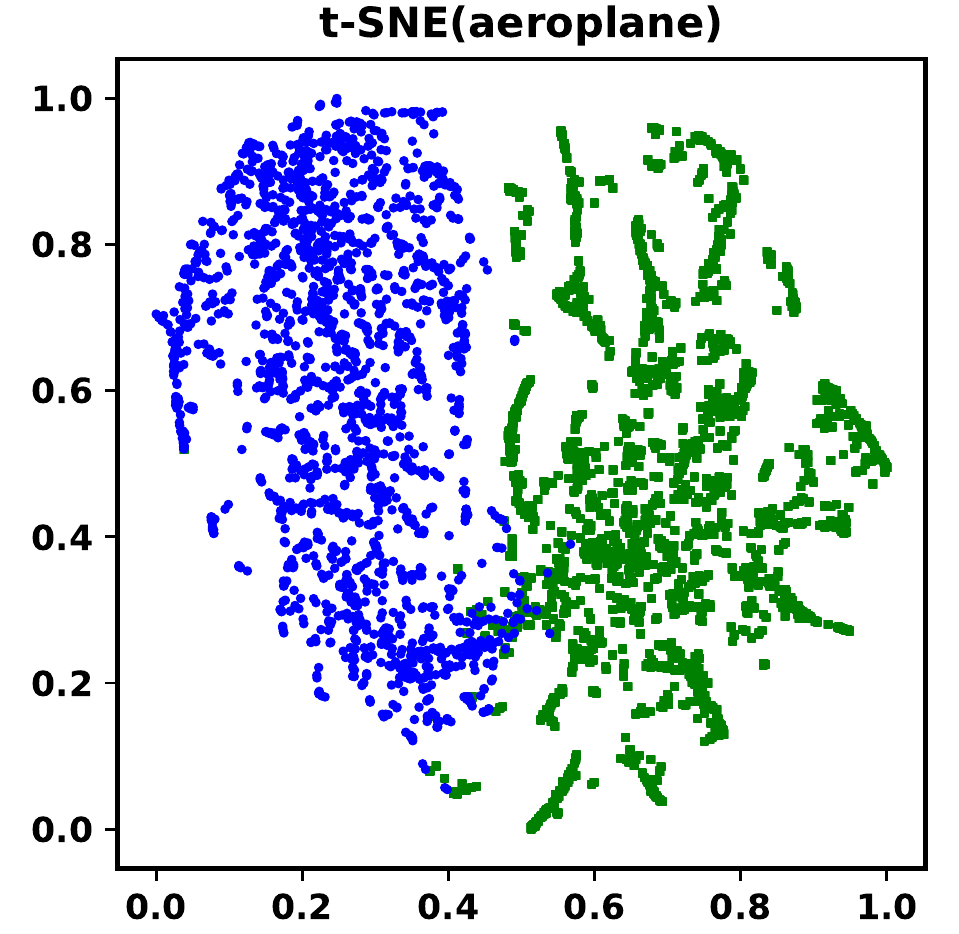}}
    \subfloat{
    \label{ssen2}
    \includegraphics[width=43mm,height=35mm]{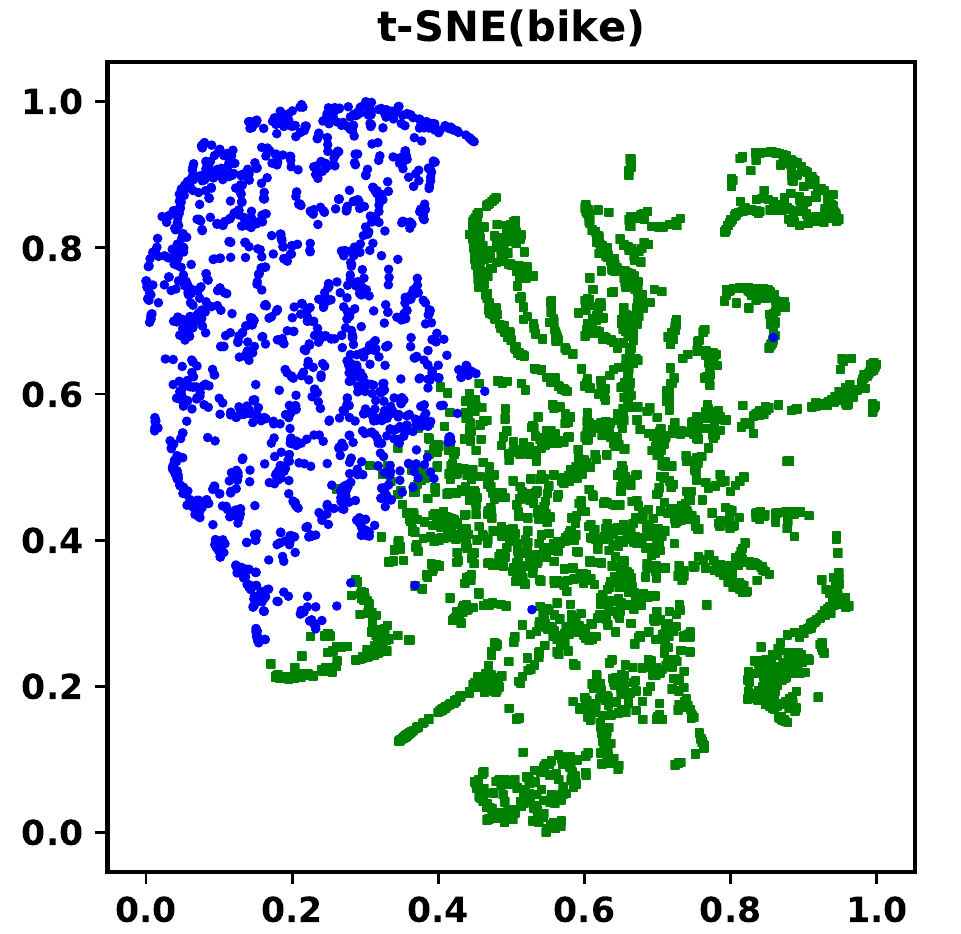}}
    \subfloat{
    \label{ssen1_}
    \includegraphics[width=43mm,height=35mm]{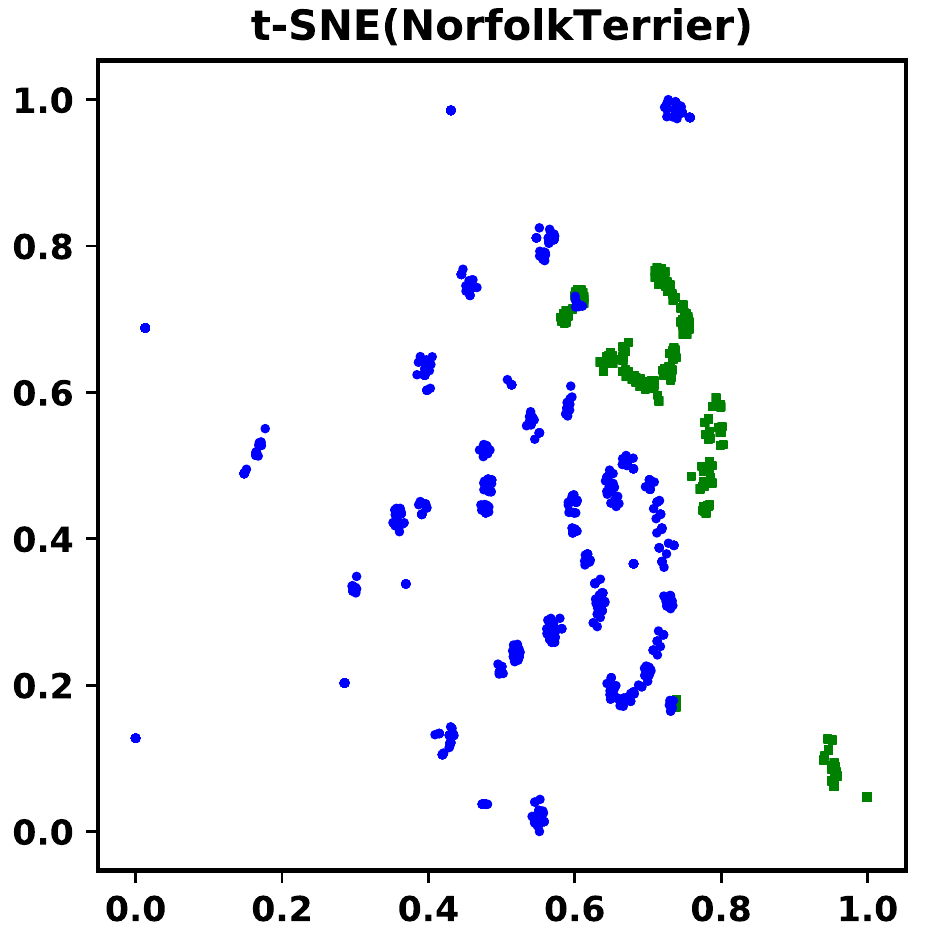}}
    \subfloat{
    \label{ssen2_}
    \includegraphics[width=43mm,height=35mm]{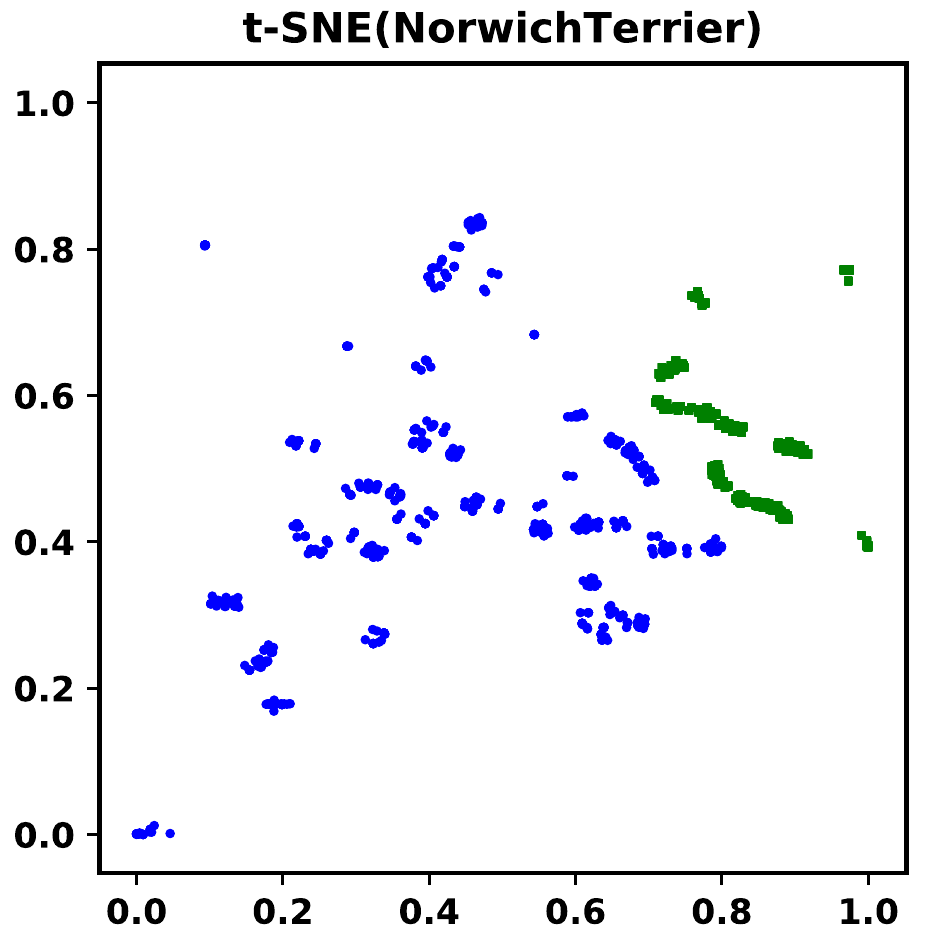}}
    \caption{Quality embedding visualization of two categories in \emph{WEB} dataset and two categories in \emph{AMT} dataset. Better distinguishability of clusters indicates better identifiability of mismatches between latent labels and noisy labels.  Blue: trustworthy embedding, Green: non-trustworthy embedding.}
    \label{fig:tsne}
\end{figure*}

\begin{table*}
\centering
\caption{Model performance (mAP) with Quantitative noise.}
\label{qNoise}
\renewcommand{\arraystretch}{1.0}
\normalsize
\scalebox{0.75}{
\begin{tabular}{c|ccccc|ccccc|ccccc}
\hline
\hline
\multicolumn{1}{l|}{} & \multicolumn{5}{c|}{\emph{V07TE}}          & \multicolumn{5}{c|}{\emph{V12TE}}          & \multicolumn{5}{c}{\emph{SD4TE}}        
\\ \hline
$P_{noise}$ & Resnet-N & LearnQ & ICNM & Bootstrap & CAN  & Resnet-N & LearnQ & ICNM & Bootstrap & CAN  & MLP-N & LearnQ & ICNM & Bootstrap & CAN  \\ \hline
1.0  & 6.4      & 9.1    & \textbf{9.2}  & 8.9       & 8.6 & 5.2      & 8.4    & 8.4  & 8.2       & \textbf{10.5} & 29.6  & 26.9   & 27.0 & 27.8      & \textbf{30.1} \\ \hline
0.8 & 33.4     & 28.0   & 28.5 & 30.1      & \textbf{36.1} & 26.6     & 23.7   & 23.8 & 25.1      & \textbf{28.0} & 41.6  & 39.6   & 39.7 & 38.6      & \textbf{49.7} \\ \hline
0.6 & 53.0     & 56.4   & 57   & 59.3      & \textbf{63.2} & 49.2     & 49.7   & 49.6 & 51.8      & \textbf{55.3} & 51.5  & 60.4   & 60.8 & 58.7      & \textbf{63.9} \\ \hline
0.4 & 70.2     & 72.0   & 71.6 & 73.3      & \textbf{79.4} & 69.0     & 70.3   & 70.5 & 72.6      & \textbf{78.4} & 73.4  & 72.7   & 73.1 & 73.5      & \textbf{77.1} \\ \hline
0.2 & 78.2     & 80.1   & 79.6 & 81.0      & \textbf{83.6} & 80.0     & 81.3   & 81.4 & 82.2      & \textbf{84.5} & 86.1  & 89.0   & 89.2 & 89.3      & \textbf{91.1} \\ \hline
0.0 & \textbf{86.8}     & 85.4   & 85.4 & 85.5      & 85.3 & \textbf{89.7}     & 88.3   & 88.3 & 88.5      & 87.3 & \textbf{96.4}  & 95.9   & 95.8 & 96.2      & 94.3 \\ \hline
\end{tabular}}
\end{table*}

\subsubsection{Impact of training size}
To explore the reliability of the proposed method when the training size changes, we compare CAN with other methods on different scales of datasets. We randomly sample different ratios of subsets in \emph{WEB} and \emph{AMT} datasets for training, and illustrate results of all the methods on \emph{V07TE}, \emph{V12TE} and \emph{SD4TE} in Fig. \ref{fig:trainSize}. 

From Fig. \ref{fig:trainSize}, the results of all methods on these datasets decline with the decrease of the training size. However, CAN performs better than other models persistently. For instance, in the left panel of Fig. \ref{fig:trainSize}, when the training size accounts at 20\%, CAN achieves 81.0$\%$ mAP on the \emph{V07TE} dataset, while ICNM and LearnQ are even worse than the most simple Resnet-N (79.4\% mAP). Similar clues can be found in the middle and right panels. These results demonstrate the reliability of CAN on different scales of datasets. 

In Fig. \ref{fig:trainSize}, we also find the decline trend on \emph{SD4TE} dataset is more significant than that on \emph{V07TE} and \emph{V12TE} datasets. This is because that even if the 20\% subset, there are still about 20k samples for training in \emph{WEB} dataset. But there are only about 1.6k samples remaining in \emph{AMT} dataset, which may lack enough knowledge to learn the classifier in the training.

\subsubsection{hyperparameter sensitivity}
To investigate the reliability of CAN with different the regularizer coefficients, we set $\lambda$ to 0, 0.2, 0.5, 1, 5, 10 to respectively validate its effect. The results are illustrated in Table. \ref{hyperparameter}. From this table, we find the performance on all datasets first grows to a peak and then gradually decreases with $\lambda$ increasing. For example, CAN achieves 85.2$\%$ mAP on \emph{V12TE} dataset when $\lambda$=0.2, but significantly decreases to below 76.6$\%$ mAP when $\lambda$=10. This indicates: (1) the regularizer in the proper degree encourages our model to find a good solution; (2) too strong regularization may induce the solution to depart from the optimal. Empirically, setting $\lambda$ between 0 to 1 makes the variational mutual information regularizer collaborate well with KL-divergences.

\subsubsection{controlled experiments with artificial noise}
In previous sections, all models are trained on \emph{WEB} and \emph{AMT} with given noise, which does not exhibit the characteristics in different noise levels. To show the superiority of CAN, we quantitatively add noise on \emph{V07TR}, \emph{V12TR} and \emph{SD4TR} datasets for training, and then compare the classification performance of all models on the \emph{V07TE}, \emph{V12TE} and \emph{SD4TE} datasets. The way to add noise to datasets is by setting a corruption probability P$_{noise}$ to randomly decide whether to shuffle elements of each clean label vector or not. We list the model performance in different P$_{noise}$ settings in Table. \ref{qNoise}.

As shown in Table. \ref{qNoise}, when the corruption probability P$_{Noise}$=1.0, the classification results of all models are close to randomness. With P$_{Noise}$ varying from 1.0 to 0, all models show improvement, since there are some clean samples available for training. Specially when P$_{Noise}$ is set to 0.8, 0.6, 0.4, 0.2, CAN robustly outperforms other baselines. However, when the training data becomes purely clean, i.e., P$_{Noise}$=0, all noise-aware models are worse than Resnet-N and MLP-N. Table. \ref{qNoise} indicates: (1) The performance of all existing models is strongly-related to the noise level in the datasets. All noise-aware models perform bad in the heavy noise. (2) When the training data is clean, noise-aware models may be worse than models without considering noise. (3) CAN shows advantages in different noise levels compared with existing methods. 

\begin{figure*}
    \centering
    \subfloat{
    \includegraphics[width=178mm]{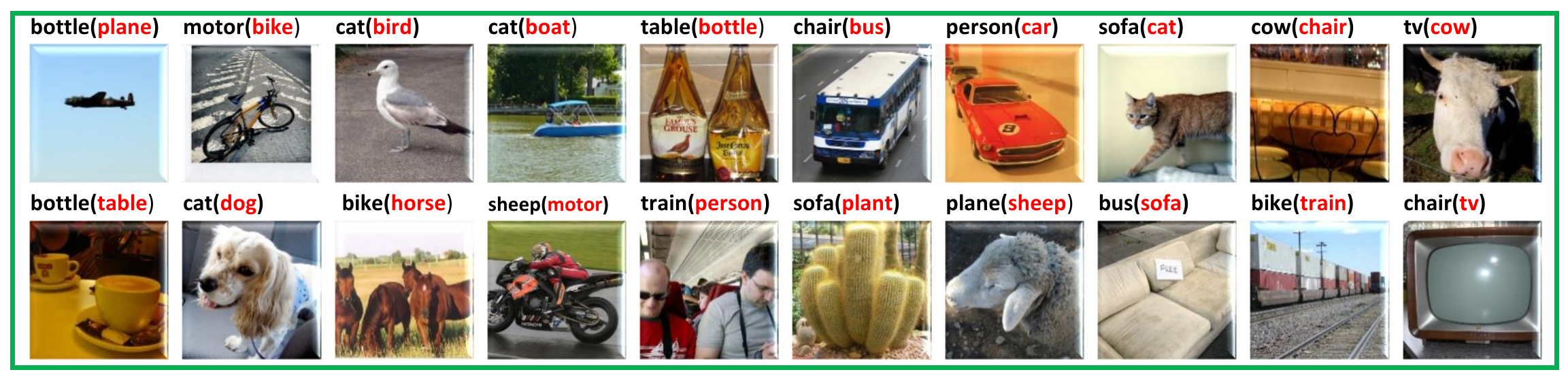}} \\
    \subfloat{
    \includegraphics[width=178mm]{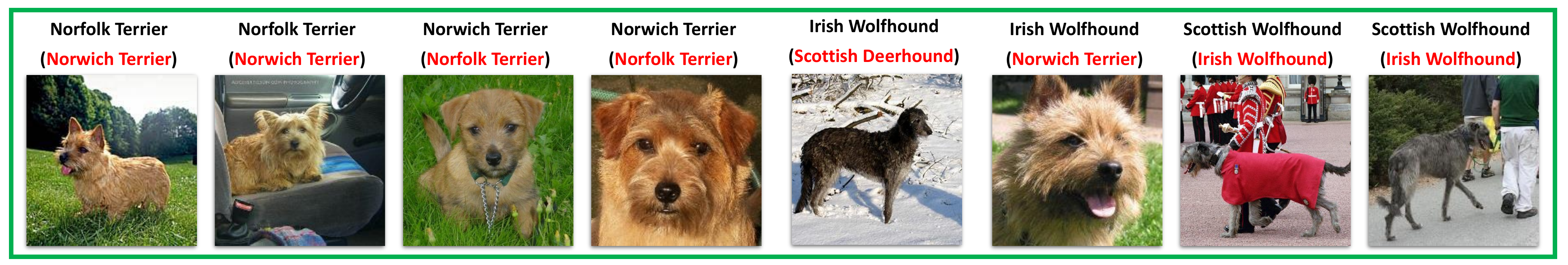}}
    \caption{\small Exemplars on latent label estimation of \emph{WEB} dataset (the first two rows) and \emph{AMT} dataset (the third row). We forward the noisy label (black word in title) and the image into CAN and compute the latent label (red word in title).}
    \label{fig:plabel}
\end{figure*}

\subsection{Model Visualization}
To give a deep insight on how CAN works, in this section, we will present the qualitative analysis about quality embedding, latent label estimation and noise transition in CAN.
\subsubsection{quality embedding}
The quality variable is estimated in the embedding space by the contrastive layer. To visualize this mechanism, we respectively forward all the training samples into CAN to compute their quality embedding. By comparing the consistency between the prior prediction (thresholded by 0.5) and the noisy label, we then binarize each embedding as {\bf trustworthy embedding} or {\bf non-trustworthy embedding}. If we only consider the Gaussian mean of each quality variable plus the embedding type, a low dimensional visualization of quality embedding can be illustrated with t-SNE package \cite{van2014accelerating}. 

In Fig. \ref{fig:tsne}, two exemplar categories ``aeroplane" and ``bike" in \emph{WEB} dataset, and two exemplar categories ``Norfolk Terrier" and ``Norwich Terrier" in \emph{AMT} dataset, are presented. As shown in Fig. \ref{fig:tsne}, the embedding in each category exhibits two distinguishable clusters. It indicates CAN can identify mismatches between latent labels and noisy labels, and selectively embed the quality variable to different subspace based on the training samples. Thus the label noise can be effectively reduced with the auxiliary of the quality variable.

Besides, we find the embedding for the first two categories are better than that for the last two categories in Fig.\ref{fig:tsne}. It is because the categories in \emph{WEB} and \emph{AMT} datasets are notably different in number and diversity of training samples. For example, there are about 4,200 different images and annotations in the ``aeroplane", while there are only about 200 different images and 1,300 annotations in the ``Norfolk Terrier". Thus embedding in the first two categories is uniformly distributed but in the last two categories is discretely cluttered.

\begin{figure*}[t]
    \centering
    \subfloat{
    \includegraphics[width=41mm]{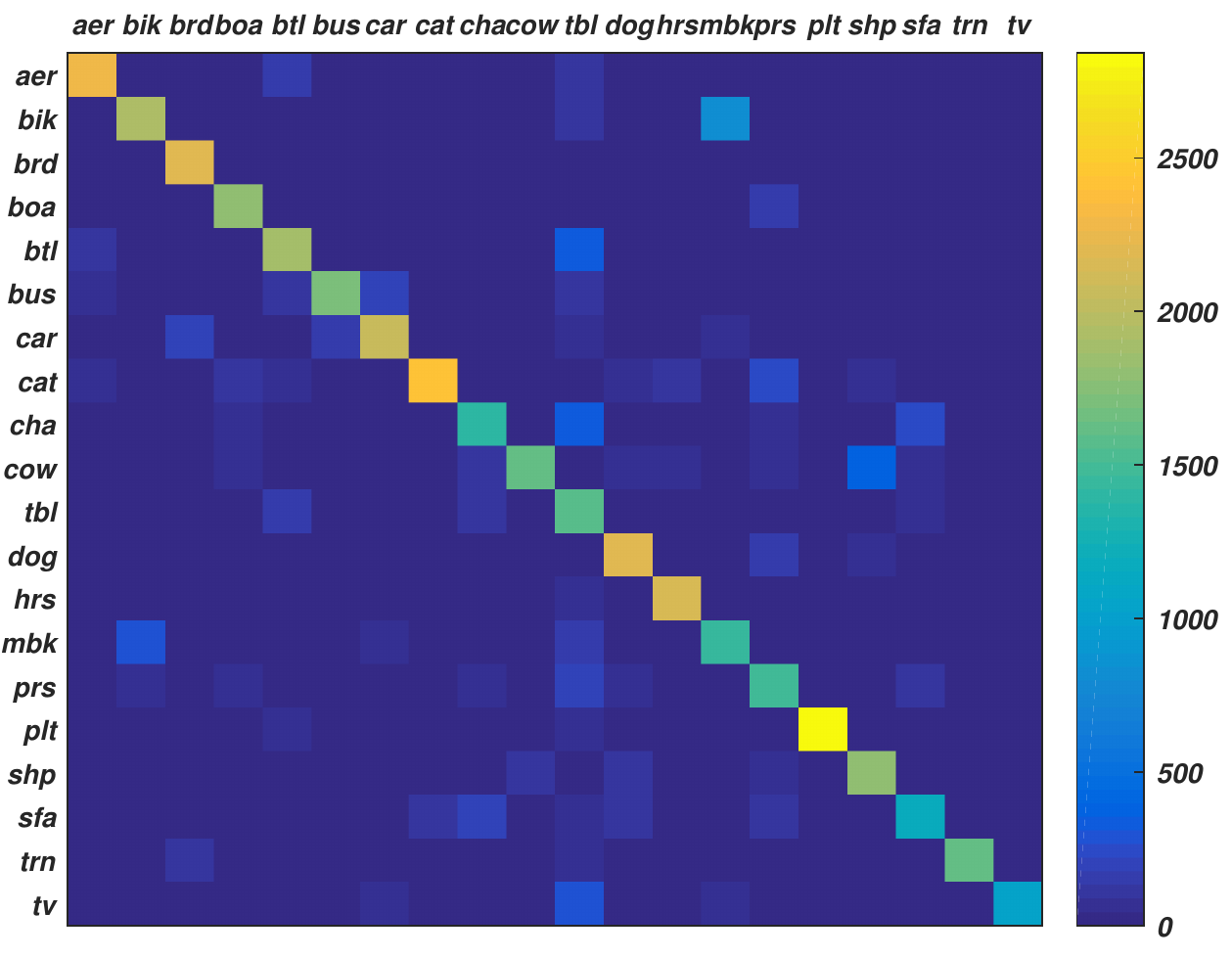}}
    \subfloat{
    \includegraphics[width=41mm]{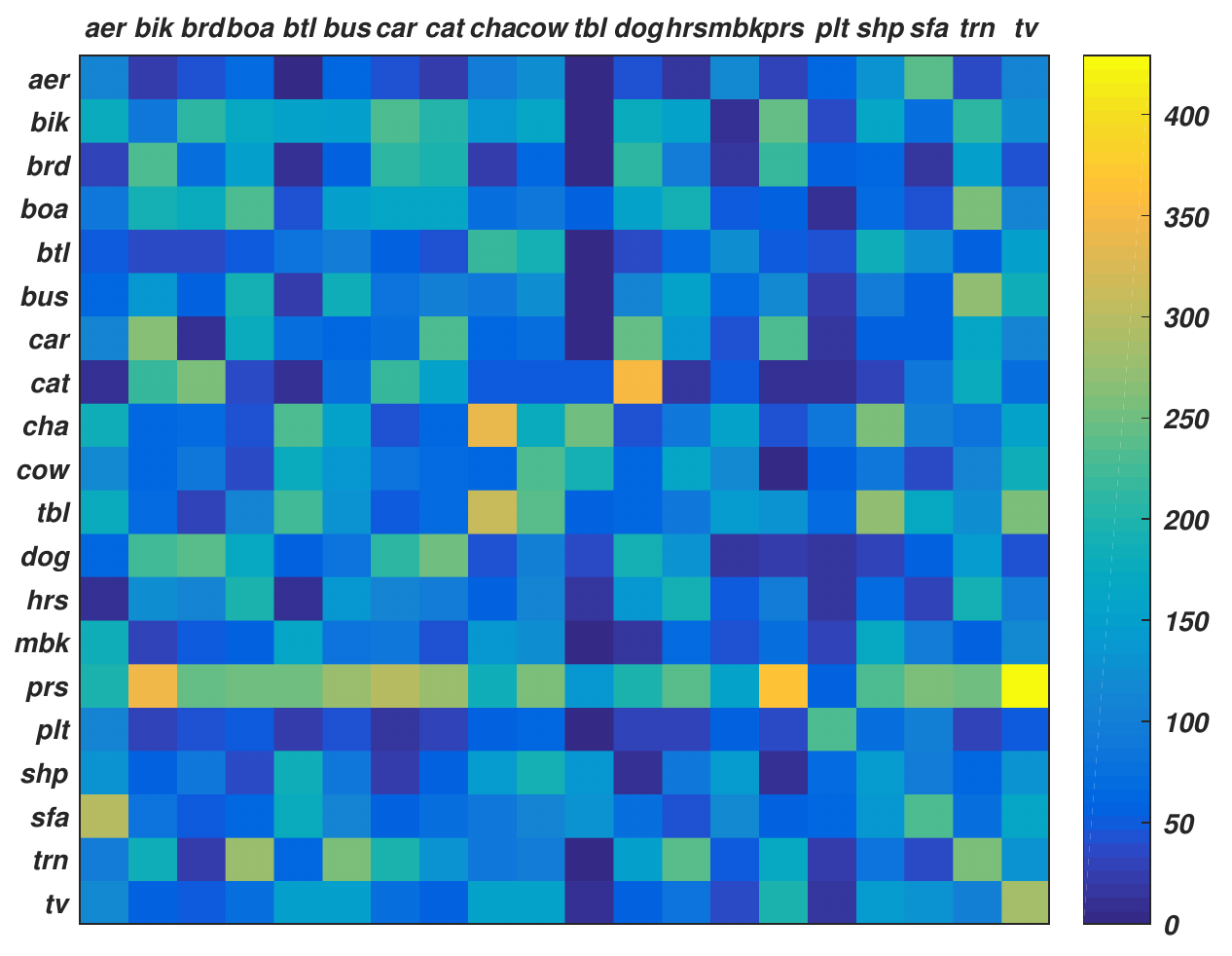}}
    \subfloat{
    \includegraphics[width=43mm]{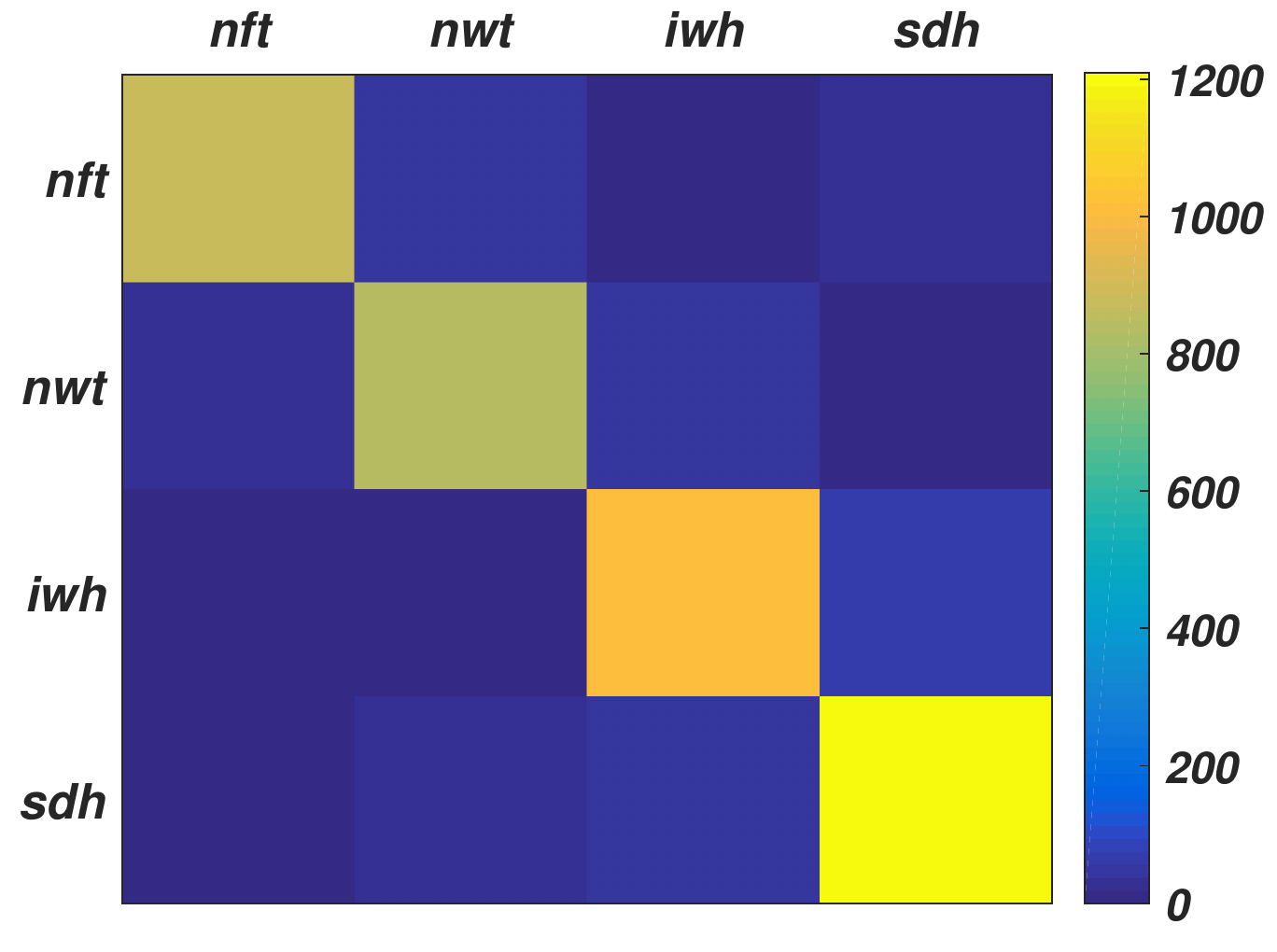}}
    \subfloat{
    \includegraphics[width=43mm]{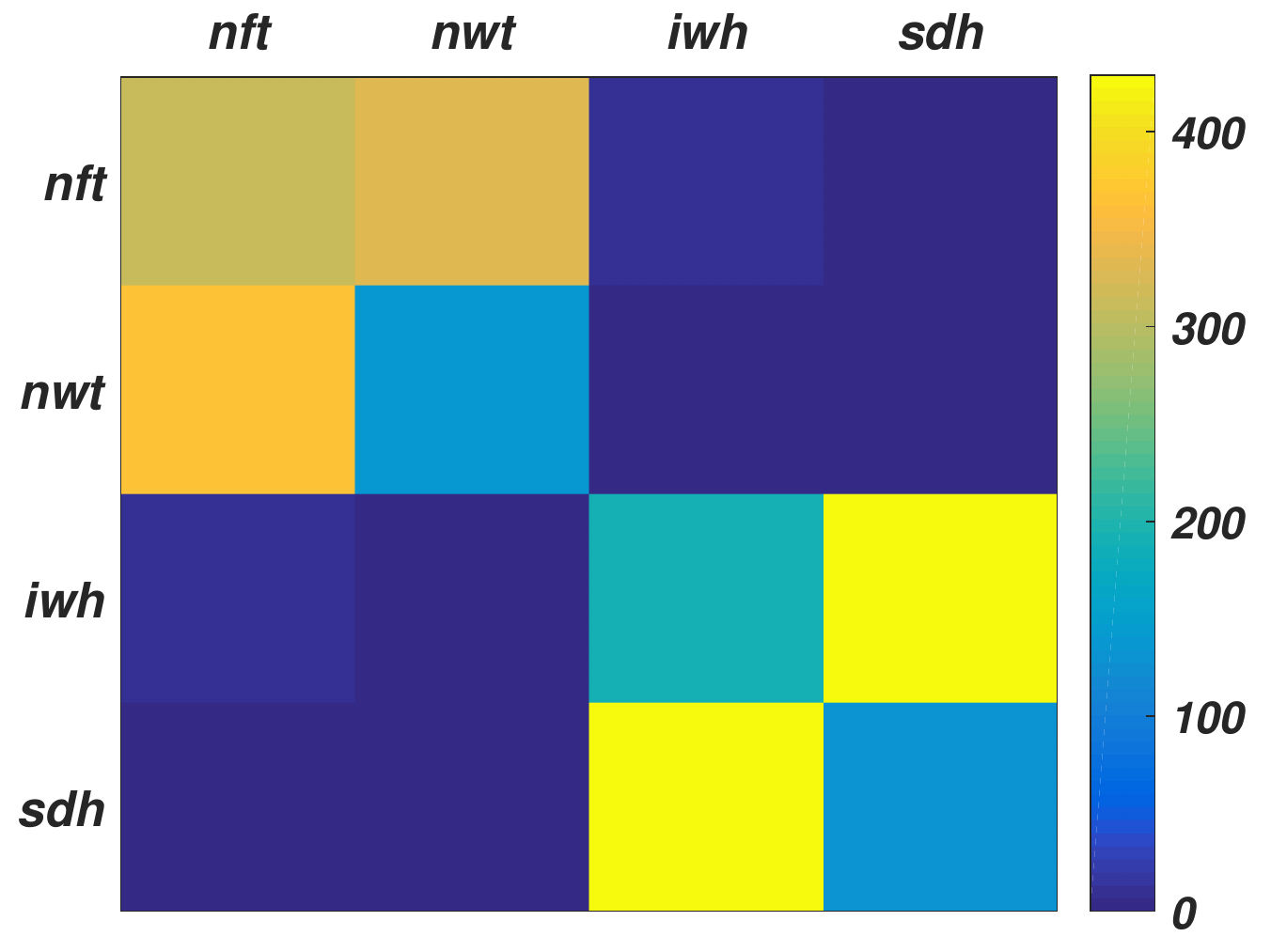}}
    \caption{\small Transition patterns among labels conditioned on trustworthy embedding and non-trustworthy embedding on \emph{WEB} dataset (the first two panels) and \emph{AMT} dataset (the last two panels). Transition conditioned on trustworthy embedding requires the consistency between the latent label and the noisy label, and thus concentrates on the diagonal. Transition conditioned on non-trustworthy embedding identifies the mismatch between the latent label and the noisy label, and thus diffuses from the diagonal. }
    \label{fig:transition}
\end{figure*}

\subsubsection{latent label estimation}
The latent label is estimated in the posterior perspective by the additive layer. To visualize this estimation, we forward all the training samples into CAN to compute output of the additive layer. In Fig. \ref{fig:plabel}, we present 20 examples of \emph{WEB} dataset and 8 examples of \emph{AMT} dataset.  

From Fig. \ref{fig:plabel}, we observe: (1) the annotations in \emph{WEB} dataset may be totally unrelated to the image content, e.g., ``bottle" for the first aeroplane image; (2) In \emph{AMT}, the Turkers also assign the wrong labels to the fine-grained images.
The former error is usually from the batch annotation function provided by the Flickr website. The latter error is usually from the limit domain knowledge of Turkers. Nevertheless, from the estimation, we find our additive layer still successfully rectifies the wrong labels. Thus based on these latent labels for training, CAN achieves the better performance than other baselines. 

\subsubsection{noise transition}
To explore how the quality embedding intermediates the mismatch between latent labels and noisy labels, we investigate the transition patterns between latent labels and noisy labels. Firstly, we forward all the training samples to CAN to compute quality embeddings and latent labels. Secondly, we utilize K-means to binarize quality embeddings (only consider Gaussian mean) into trustworthy embedding and non-trustworthy embedding. Thirdly, we count transitions from latent labels to noisy labels conditioned on two types of embeddings. In Fig. \ref{fig:transition}, we respectively plot two transition patterns with heatmaps for \emph{WEB} dataset and \emph{AMT} dataset. 

As shown for \emph{WEB} dataset in Fig. \ref{fig:transition}, the diagonal of the transition pattern conditioned on trustworthy embedding is dominant. In this case, noisy labels are considered to be reliable and thus transition should mainly happen among same labels. However, the transition patterns conditioned on non-trustworthy embedding is diffusing. Because in this case, noisy labels are considered not correct and transition usually happen between different labels. Similarly, transition patterns on \emph{AMT} dataset in Fig. \ref{fig:transition} also have these characteristics. Fig. \ref{fig:transition} indicates CAN is based on quality embedding to automatically disturb the latent label to match the noisy label.

The transition pattern conditioned on non-trustworthy embedding usually reflects the real-world noise. Some interesting patterns can be found. For instance, according to the second panel of Fig. \ref{fig:transition}, ``plt'' class has less transition to other classes while the transition between ``prs" and ``tv" has high value. It means: (1) people who upload the ``pottedplant" images to social websites almost do not annotate it wrong; (2) for ``tv" images, some people focus on persons in the TV program, and others may pay attention to TV itself. Similarly in the fourth panel of Fig. \ref{fig:transition}, the transition on \emph{AMT} usually exists in the appear-similar dogs, i.e., ``Norfolk Terrier" and ``Norwich Terrier", ``Irish Wolfhound" and ``Scottish Wolfhound". It reflects that it is more difficult to distinguish these two breeds of dogs than other pairs in some sense.

\section{Conclusion}
In this paper, we present a quality embedding model to learn the classifier from noisy image labels, which effectively avoid the error back-propagated from label noise. To instantiate the model, a Contrastive-Additive Noise network is well-designed. Regarding parameter estimation, we deduce an efficient SGD optimization algorithm by applying recent discrete and continuous reparameterization tricks. We demonstrate our model outperform other noise-aware deep learning methods on some noisy training datasets. Simultaneously, detailed visualization on three key parts is presented to give a deep insight on our model. However, we only validate our model in image data in this paper and other types of contents can be further explored.


%

\appendices

\section{Computation for KL-divergences and regularizers}
The remaining four terms in the RHS of Eq. \eqref{objective} can be calculated without sampling. For example, for the latent label $z_m$, both  $q(z_m|x_m,y_m)$ and $P(z_m|x_m)$ are two $K$-dimensional multinomial probabilities. Their KL-divergence term and regularizer can be simplified by enumerating each dimension. For the quality variable $s_m$, it is from the $D$-dimensional Gaussian space $ N(\mu(x_m,y_m),\text{diag}(\sigma^2(x_m,y_m)))$ whose parameters are implicitly modeled with network of input $x_m$ and $y_m$. If we assume its prior $P(s_m)$ is $N(\mathbf{0},\mathbf{1})$ like \cite{kingma2014auto}, it is easy to compute their KL-divergence and the regularizer due to the conjugation. In Eq. \eqref{rkl}, we give their simplifications bigeminally.
\begin{align}\label{rkl}
&\mathbf{D_{KL}}\left[q(z_m|x_m,y_m)||P(z_m|x_m)\right] - \lambda \mathbf{E}_{q(z_m|x_m,y_m)}\left[\ln{q(z_m|x_m,y_m)}\right] \nonumber \\
& = \sum_{k=1}^K \sum_{z_{mk}} q(z_{mk}|x_m,y_m)\ln  \frac{q(z_{mk}|x_m,y_m)^{1-\lambda}}{P(z_{mk}|x_m)}\nonumber \\
    & \mathbf{D_{KL}}\left[ q(s_m|x_m,y_m)|| P(s_m)\right] -  \lambda \mathbf{E}_{q(s_m|x_m,y_m)}\left[\ln{q(s_m|x_m,y_m)}\right] \nonumber \\
& = - \frac{1}{2} \sum_{d=1}^{D} \left( (1-\lambda)\ln \sigma_d^2(x_m,y_m) - \sigma_d^2(x_m,y_m)\right) \nonumber \\
&\quad + \frac{1}{2} \mu(x_m,y_m)^T\mu(x_m,y_m) + const  
\end{align}

\ifCLASSOPTIONcompsoc
  \section*{Acknowledgments}
\else
  \section*{Acknowledgment}
\fi

\ifCLASSOPTIONcaptionsoff
  \newpage
\fi



\bibliographystyle{IEEEtran}
%
\bibliography{reference} 

\begin{thebibliography}{10}
\providecommand{\url}[1]{#1}
\csname url@samestyle\endcsname
\providecommand{\newblock}{\relax}
\providecommand{\bibinfo}[2]{#2}
\providecommand{\BIBentrySTDinterwordspacing}{\spaceskip=0pt\relax}
\providecommand{\BIBentryALTinterwordstretchfactor}{4}
\providecommand{\BIBentryALTinterwordspacing}{\spaceskip=\fontdimen2\font plus
\BIBentryALTinterwordstretchfactor\fontdimen3\font minus
  \fontdimen4\font\relax}
\providecommand{\BIBforeignlanguage}[2]{{%
\expandafter\ifx\csname l@#1\endcsname\relax
\typeout{** WARNING: IEEEtran.bst: No hyphenation pattern has been}%
\typeout{** loaded for the language `#1'. Using the pattern for}%
\typeout{** the default language instead.}%
\else
\language=\csname l@#1\endcsname
\fi
#2}}
\providecommand{\BIBdecl}{\relax}
\BIBdecl

\bibitem{NIPS2012_4824}
A.~Krizhevsky, I.~Sutskever, and G.~E. Hinton, ``Imagenet classification with
  deep convolutional neural networks,'' in \emph{Advances in Neural Information
  Processing Systems 25}, F.~Pereira, C.~J.~C. Burges, L.~Bottou, and K.~Q.
  Weinberger, Eds.\hskip 1em plus 0.5em minus 0.4em\relax Curran Associates,
  Inc., 2012, pp. 1097--1105.

\bibitem{simonyan2014very}
K.~Simonyan and A.~Zisserman, ``Very deep convolutional networks for
  large-scale image recognition,'' 2014.

\bibitem{Szegedy_2015_CVPR}
C.~Szegedy, W.~Liu, Y.~Jia, P.~Sermanet, S.~Reed, D.~Anguelov, D.~Erhan,
  V.~Vanhoucke, and A.~Rabinovich, ``Going deeper with convolutions,'' in
  \emph{The IEEE Conference on Computer Vision and Pattern Recognition (CVPR)},
  June 2015.

\bibitem{he2016deep}
K.~He, X.~Zhang, S.~Ren, and J.~Sun, ``Deep residual learning for image
  recognition,'' in \emph{Proceedings of the IEEE Conference on Computer Vision
  and Pattern Recognition}, 2016, pp. 770--778.

\bibitem{wang2014weakly}
C.~Wang, W.~Ren, K.~Huang, and T.~Tan, ``Weakly supervised object localization
  with latent category learning,'' in \emph{European Conference on Computer
  Vision}.\hskip 1em plus 0.5em minus 0.4em\relax Springer, 2014, pp. 431--445.

\bibitem{bilen2016weakly}
H.~Bilen and A.~Vedaldi, ``Weakly supervised deep detection networks,'' in
  \emph{Proceedings of the IEEE Conference on Computer Vision and Pattern
  Recognition}, 2016, pp. 2846--2854.

\bibitem{wang2014joint}
L.~Wang, G.~Hua, J.~Xue, Z.~Gao, and N.~Zheng, ``Joint segmentation and
  recognition of categorized objects from noisy web image collection,''
  \emph{IEEE Transactions on Image Processing}, vol.~23, no.~9, pp. 4070--4086,
  2014.

\bibitem{Zhang_2015_CVPR}
W.~Zhang, S.~Zeng, D.~Wang, and X.~Xue, ``Weakly supervised semantic
  segmentation for social images,'' in \emph{The IEEE Conference on Computer
  Vision and Pattern Recognition (CVPR)}, June 2015.

\bibitem{Khoreva_2017_CVPR}
A.~Khoreva, R.~Benenson, J.~Hosang, M.~Hein, and B.~Schiele, ``Simple does it:
  Weakly supervised instance and semantic segmentation,'' in \emph{The IEEE
  Conference on Computer Vision and Pattern Recognition (CVPR)}, July 2017.

\bibitem{lu2017learning}
Z.~Lu, Z.~Fu, T.~Xiang, P.~Han, L.~Wang, and X.~Gao, ``Learning from weak and
  noisy labels for semantic segmentation,'' \emph{IEEE transactions on pattern
  analysis and machine intelligence}, vol.~39, no.~3, pp. 486--500, 2017.

\bibitem{Divvala_2014_CVPR}
S.~K. Divvala, A.~Farhadi, and C.~Guestrin, ``Learning everything about
  anything: Webly-supervised visual concept learning,'' in \emph{The IEEE
  Conference on Computer Vision and Pattern Recognition (CVPR)}, June 2014.

\bibitem{chen2015webly}
X.~Chen and A.~Gupta, ``Webly supervised learning of convolutional networks,''
  in \emph{2015 IEEE International Conference on Computer Vision (ICCV)}, 2015,
  pp. 1431--1439.

\bibitem{krishna2017visual}
R.~Krishna, Y.~Zhu, O.~Groth, J.~Johnson, K.~Hata, J.~Kravitz, S.~Chen,
  Y.~Kalantidis, L.-J. Li, D.~A. Shamma \emph{et~al.}, ``Visual genome:
  Connecting language and vision using crowdsourced dense image annotations,''
  \emph{International Journal of Computer Vision}, vol. 123, no.~1, pp. 32--73,
  2017.

\bibitem{Sukhbaatar2015Training}
S.~Sukhbaatar, J.~Bruna, M.~Paluri, L.~Bourdev, and R.~Fergus, ``Training
  convolutional networks with noisy labels,'' \emph{Computer Science}, 2015.

\bibitem{raykar2010learning}
V.~C. Raykar, S.~Yu, L.~H. Zhao, G.~H. Valadez, C.~Florin, L.~Bogoni, and
  L.~Moy, ``Learning from crowds,'' \emph{Journal of Machine Learning
  Research}, vol.~11, no. Apr, pp. 1297--1322, 2010.

\bibitem{natarajan2013learning}
N.~Natarajan, I.~S. Dhillon, P.~Ravikumar, and A.~Tewari, ``Learning with noisy
  labels,'' \emph{Advances in Neural Information Processing Systems}, vol.~26,
  pp. 1196--1204, 2013.

\bibitem{Liu2014Classification}
T.~Liu and D.~Tao, ``Classification with noisy labels by importance
  reweighting,'' \emph{IEEE Transactions on Pattern Analysis \& Machine
  Intelligence}, vol.~38, no.~3, p. 447, 2014.

\bibitem{zhou2012learning}
D.~Zhou, S.~Basu, Y.~Mao, and J.~C. Platt, ``Learning from the wisdom of crowds
  by minimax entropy,'' in \emph{Advances in Neural Information Processing
  Systems}, 2012, pp. 2195--2203.

\bibitem{Frenay2014classification}
B.~Frenay and M.~Verleysen, ``Classification in the presence of label noise: A
  survey,'' \emph{IEEE Transactions on Neural Networks and Learning Systems},
  vol.~25, no.~5, pp. 845--869, May 2014.

\bibitem{Mnih2012Learning}
V.~Mnih and G.~Hinton, ``Learning to label aerial images from noisy data,'' in
  \emph{International Conference on Machine Learning}, 2012.

\bibitem{azadi2015auxiliary}
S.~Azadi, J.~Feng, S.~Jegelka, and T.~Darrell, ``Auxiliary image regularization
  for deep cnns with noisy labels,'' 2016.

\bibitem{misra2016seeing}
I.~Misra, C.~Lawrence~Zitnick, M.~Mitchell, and R.~Girshick, ``Seeing through
  the human reporting bias: Visual classifiers from noisy human-centric
  labels,'' in \emph{Proceedings of the IEEE Conference on Computer Vision and
  Pattern Recognition}, 2016, pp. 2930--2939.

\bibitem{Izadinia2015Deep}
H.~Izadinia, B.~C. Russell, A.~Farhadi, M.~D. Hoffman, and A.~Hertzmann, ``Deep
  classifiers from image tags in the wild,'' in \emph{The Workshop on
  Community-Organized Multimodal Mining: Opportunities for Novel Solutions},
  2015, pp. 13--18.

\bibitem{reed2014training}
S.~Reed, H.~Lee, D.~Anguelov, C.~Szegedy, D.~Erhan, and A.~Rabinovich,
  ``Training deep neural networks on noisy labels with bootstrapping,''
  \emph{Computer Science}, 2014.

\bibitem{patrini2016making}
G.~Patrini, A.~Rozza, A.~Menon, R.~Nock, and L.~Qu, ``Making neural networks
  robust to label noise: a loss correction approach,'' \emph{arXiv preprint
  arXiv:1609.03683}, 2016.

\bibitem{xiao2015learning}
T.~Xiao, T.~Xia, Y.~Yang, C.~Huang, and X.~Wang, ``Learning from massive noisy
  labeled data for image classification,'' in \emph{Proceedings of the IEEE
  Conference on Computer Vision and Pattern Recognition}, 2015, pp. 2691--2699.

\bibitem{jindal2016learning}
I.~Jindal, M.~Nokleby, and X.~Chen, ``Learning deep networks from noisy labels
  with dropout regularization,'' in \emph{Data Mining (ICDM), 2016 IEEE 16th
  International Conference on}.\hskip 1em plus 0.5em minus 0.4em\relax IEEE,
  2016, pp. 967--972.

\bibitem{li2017learning}
Y.~Li, J.~Yang, Y.~Song, L.~Cao, J.~Luo, and J.~Li, ``Learning from noisy
  labels with distillation,'' \emph{arXiv preprint arXiv:1703.02391}, 2017.

\bibitem{veit17learning}
A.~Veit, N.~Alldrin, G.~Chechik, I.~Krasin, A.~Gupta, and S.~Belongie,
  ``Learning from noisy large-scale datasets with minimal supervision,''
  \emph{Computer Vision and Pattern Recognition (CVPR)}, 2017.

\bibitem{wainwright2008graphical}
M.~J. Wainwright, M.~I. Jordan \emph{et~al.}, ``Graphical models, exponential
  families, and variational inference,'' \emph{Foundations and
  Trends{\textregistered} in Machine Learning}, vol.~1, no. 1--2, pp. 1--305,
  2008.

\bibitem{ICML2012Paisley_687}
D.~M. Blei, M.~I. Jordan, and J.~W. Paisley, ``Variational bayesian inference
  with stochastic search,'' in \emph{Proceedings of the 29th International
  Conference on Machine Learning (ICML-12)}, J.~Langford and J.~Pineau,
  Eds.\hskip 1em plus 0.5em minus 0.4em\relax New York, NY, USA: ACM, 2012, pp.
  1367--1374.

\bibitem{blei2017variational}
D.~M. Blei, A.~Kucukelbir, and J.~D. McAuliffe, ``Variational inference: A
  review for statisticians,'' \emph{Journal of the American Statistical
  Association}, no. just-accepted, 2017.

\bibitem{nettleton2010}
D.~F. Nettleton, A.~Orriols-Puig, and A.~Fornells, ``A study of the effect of
  different types of noise on the precision of supervised learning
  techniques,'' \emph{Artificial Intelligence Review}, vol.~33, no.~4, pp.
  275--306, 2010.

\bibitem{Joulin2015Learning}
A.~Joulin, L.~V.~D. Maaten, A.~Jabri, and N.~Vasilache, \emph{Learning Visual
  Features from Large Weakly Supervised Data}.\hskip 1em plus 0.5em minus
  0.4em\relax Springer International Publishing, 2015.

\bibitem{ganchev2010posterior}
K.~Ganchev, J.~Gillenwater, B.~Taskar \emph{et~al.}, ``Posterior regularization
  for structured latent variable models,'' \emph{Journal of Machine Learning
  Research}, vol.~11, no. Jul, pp. 2001--2049, 2010.

\bibitem{NIPS2010_4154}
A.~Krause, P.~Perona, and R.~G. Gomes, ``Discriminative clustering by
  regularized information maximization,'' in \emph{Advances in Neural
  Information Processing Systems 23}, J.~D. Lafferty, C.~K.~I. Williams,
  J.~Shawe-Taylor, R.~S. Zemel, and A.~Culotta, Eds.\hskip 1em plus 0.5em minus
  0.4em\relax Curran Associates, Inc., 2010, pp. 775--783.

\bibitem{zhu2014bayesian}
J.~Zhu, N.~Chen, and E.~P. Xing, ``Bayesian inference with posterior
  regularization and applications to infinite latent svms,'' \emph{Journal of
  Machine Learning Research}, vol.~15, p. 1799, 2014.

\bibitem{NIPS2016_6399}
X.~Chen, X.~Chen, Y.~Duan, R.~Houthooft, J.~Schulman, I.~Sutskever, and
  P.~Abbeel, ``Infogan: Interpretable representation learning by information
  maximizing generative adversarial nets,'' in \emph{Advances in Neural
  Information Processing Systems 29}, D.~D. Lee, M.~Sugiyama, U.~V. Luxburg,
  I.~Guyon, and R.~Garnett, Eds.\hskip 1em plus 0.5em minus 0.4em\relax Curran
  Associates, Inc., 2016, pp. 2172--2180.

\bibitem{kingma2014auto}
D.~P. Kingma and M.~Welling, ``Auto-encoding variational bayes,'' \emph{stat},
  vol. 1050, p.~10, 2014.

\bibitem{jang2016categorical}
E.~Jang, S.~Gu, and B.~Poole, ``Categorical reparameterization with
  gumbel-softmax,'' \emph{arXiv preprint arXiv:1611.01144}, 2016.

\bibitem{higgins2016beta}
I.~Higgins, L.~Matthey, A.~Pal, C.~Burgess, X.~Glorot, M.~Botvinick,
  S.~Mohamed, and A.~Lerchner, ``beta-vae: Learning basic visual concepts with
  a constrained variational framework,'' 2016.

\bibitem{van2016conditional}
A.~van~den Oord, N.~Kalchbrenner, L.~Espeholt, O.~Vinyals, A.~Graves
  \emph{et~al.}, ``Conditional image generation with pixelcnn decoders,'' in
  \emph{Advances in Neural Information Processing Systems}, 2016, pp.
  4790--4798.

\bibitem{Thomee2015The}
B.~Thomee, D.~A. Shamma, G.~Friedland, B.~Elizalde, K.~Ni, D.~Poland, D.~Borth,
  and L.~J. Li, ``The new data and new challenges in multimedia research,''
  \emph{Communications of the Acm}, vol.~59, no.~2, pp. 64--73, 2015.

\bibitem{pascal-voc-2007}
M.~Everingham, L.~Van~Gool, C.~K.~I. Williams, J.~Winn, and A.~Zisserman, ``The
  pascal visual object classes challenge 2007 results,'' 2007.

\bibitem{KhoslaYaoJayadevaprakashFeiFei_FGVC2011}
A.~Khosla, N.~Jayadevaprakash, B.~Yao, and L.~Fei-Fei, ``Novel dataset for
  fine-grained image categorization,'' in \emph{First Workshop on Fine-Grained
  Visual Categorization, IEEE Conference on Computer Vision and Pattern
  Recognition}, Colorado Springs, CO, June 2011.

\bibitem{pascal-voc-2012}
M.~Everingham, L.~Van~Gool, C.~K.~I. Williams, J.~Winn, and A.~Zisserman, ``The
  pascal visual object classes challenge 2012 results,'' 2012.

\bibitem{van2014accelerating}
L.~Van Der~Maaten, ``Accelerating t-sne using tree-based algorithms.''
  \emph{Journal of machine learning research}, vol.~15, no.~1, pp. 3221--3245,
  2014.

\end{thebibliography}

%

\begin{IEEEbiography}{}
\end{IEEEbiography}

\begin{IEEEbiographynophoto}{}
\end{IEEEbiographynophoto}


\begin{IEEEbiographynophoto}{}
\end{IEEEbiographynophoto}




\end{document}